# A Review Paper of the Effects of Distinct Modalities and ML Techniques to Distracted Driving Detection

Anthony. Dontoh, *Member, IEEE*, Stephanie. Ivey, Logan. Sirbaugh, *Member, IEEE* and Armstrong. Aboah, *Member, IEEE*

**Abstract** Distracted driving remains a significant global challenge with severe human and economic repercussions, demanding improved detection and intervention strategies. While previous studies have extensively explored single-modality approaches, recent research indicates that these systems often fall short in identifying complex distraction patterns, particularly cognitive distractions. This systematic review addresses critical gaps by providing a comprehensive analysis of machine learning (ML) and deep learning (DL) techniques applied across various data modalities—visual, sensory, auditory, and multimodal. By categorizing and evaluating studies based on modality, data accessibility, and methodology, this review clarifies which approaches yield the highest accuracy and are best suited for specific distracted driving detection goals. The findings offer clear guidance on the advantages of multimodal versus single-modal systems and capture the latest advancements in the field. Ultimately, this review contributes valuable insights for developing robust distracted driving detection frameworks, supporting enhanced road safety and mitigation strategies.

## I. INTRODUCTION

Road crashes are a critical global concern, with devastating human and economic consequences [1]. The World Health Organization (WHO) reports that approximately 1.19 million people lose their lives annually in road traffic accidents, with an additional 20 to 50 million sustaining non-fatal injuries, many of which lead to permanent disabilities [2]. Beyond the profound human toll, these accidents impose a significant economic burden on nations, accounting for an estimated 1 to 3 percent of Gross Domestic Product (GDP), with some countries facing costs as high as 6 percent [2]. These staggering impacts underscore the urgent need for continued efforts to enhance road safety and reduce the incidence of road crashes.

A crucial step in enhancing road safety is to understand the root causes of accidents. The National Highway Traffic Safety Administration (NHTSA) identifies distracted driving as one of the leading causes of road accidents [3]–[5]. For instance, the NHTSA reported in 2022 that distracted driving claimed over 3,000 lives and injured approximately 289,000 individuals in the United States alone [6]. Similarly, distracted driving accounts for 16% of fatal crashes in Australia and between 12% and 14% in Norway [7]. Beyond fatalities, distracted driving disrupts traffic efficiency, leading to erratic behaviors, traffic delays, increased fuel consumption, and exacerbated congestion

[8], [9]. These aggravated impacts of distracted driving have caught the attention of several empirical studies [10].

Distracted driving encompasses any activity that diverts drivers' attention from driving, including cognitive, manual, visual, or auditory distractions, [6] highlighting the complexity of this behavior. This complexity has driven researchers to leverage both *simulation-based datasets* and *naturalistic driving study* (NDS) data to understand and mitigate its effects. In recent years, there has been a shift towards applying *machine learning* (ML) and *deep learning* (DL) techniques across these varied datasets for a comprehensive understanding of distracted driving behaviors [11]–[14]. These advanced methods have demonstrated potential in capturing the intricate patterns of distracted driving behaviors in real-world scenarios, becoming a focal point for researchers [15].

Review papers by scholars have contributed significantly to our understanding of distracted driving behaviors, offering comprehensive overviews on various aspects of the subject. For instance, the study by Young et al. [16] focused *narrowly on in-vehicle distractions*, particularly cell phone use, while Papatheocharous et al. [15] examined *smartphone-based monitoring*, advocating for smartphone sensor-based data for distracted driving detection. Other scholarly review articles have focused on *specific aspects* such as *data collection* methods, *analysis techniques*, and *crash prevention strategies* [3], the *selection of metrics* for evaluating distracted driving countermeasures [17] and the *effectiveness of Convolutional Neural Networks* (CNNs) in real-time distraction detection [18]. Despite these valuable contributions, researchers have identified gaps in the existing literature. Lansdown et al. [19], for instance, critiqued systematic review efforts, highlighting the *need for methodological rigor*, thus the need for more comprehensive analyses.

More recent review papers have addressed some of the methodological gaps highlighted by Lansdown [10]. Wang et al. [20] primarily focused on analyzing driver behavior using visual data obtained from in-vehicle cameras, reviewing techniques for monitoring driver states such as fatigue, drowsiness, and attentiveness using facial features, eye movements and head pose. Kashevnik et al. [21] in 2021 explored various methods, including the integration of visual, sensor, and physiological data to detect driver distraction, providing a framework for how multiple data sources can enhance detection accuracy and reliability. Hassan et al. [3] broadened the scope by discussing data collection methods and crash prevention strategies, highlighting the importance of naturalistic driving data in understanding real-world behaviors.

However, limitations remain across these review studies. They fall short in providing a comparative analysis of different modalities—visual, auditory, sensory, or combined approaches—that could enhance distracted driving detection. Studies such as those by Wang et al. [20] are constrained by their focus on visual data, offering limited exploration of other data modalities. Kashevnik et al. [21] center on data collection techniques and prevention strategies but provide little comparative analysis of how various modalities perform in distraction detection. Similarly, Hassan et al. [3] offers a broad overview of data collection methods, yet primarily concentrates on crash prevention rather than a detailed assessment of how different data modalities such as visual, sensor, and physiological impact the effectiveness of distraction detection. These gaps highlight the need for a more comprehensive review that directly compares distinct modalities and multimodal approaches, assessing their strengths and limitations in enhancing distracted driving detection accuracy. Also, given the rapid evolution of techniques and approaches, there is a pressing need for an up-to-date analysis of current advances in distracted driving detection research.

Therefore, this study aims to address these gaps by identifying and categorizing various ML and DL approaches applied to visual, sensory, auditory, and multimodal data in distracted driving studies. Through an up-to-date comparative analysis, this study will highlight which approaches have proven most beneficial in specific contexts, thereby providing valuable guidance for future research in this rapidly evolving field.

The study makes three important contributions to the methodological conceptualization and application of ML and DL techniques in distractive driving. First, it will bring clarity on which modality coupled with methodological approach to distracted driving studies yields the best accuracy. Second, there will be clear guidance on whether single-modal analysis or multimodal analysis is best suited for a particular distracted driving study objective. Third, the recent studies that have made significant advancements will be captured to provide the most accurate information on distracted driving research approaches. Ultimately, this review will contribute to enhancing road safety by improving distracted driving detection approaches for the next generation of Advanced Driver Assistance Systems (ADAS) and informing more effective mitigation strategies.

This systematic review focused on peer-reviewed studies that applied machine learning (ML) and deep learning (DL) techniques to distracted driving. The study incorporated a comprehensive search of databases including IEEE Xplore, PubMed, Scopus, SpringerLink, ScienceDirect, ACM Digital Library, ASCE, TRB, and Google Scholar. These platforms were selected for their comprehensive coverage of relevant fields, including electrical engineering, computer science, traffic safety, and health sciences. This broad-based approach ensured a multidisciplinary perspective on the technologies used to detect and mitigate distracted driving, encompassing the latest advancements in computational methods and their practical applications in improving road safety.

The remainder of this paper is structured as follows: Initially, the methodology for selecting and analyzing relevant studies is presented. Next, the categorization of studies based on whether they used public or private datasets, followed by an examination of data modalities and the ML or DL techniques employed is highlighted. During this categorization, the strengths and weaknesses of the approaches were critically analyzed to provide a nuanced understanding of the current state of research, and this is discussed. A synthesis of the implications of these findings for future research and practical applications in mitigating distracted driving follows the synthesis. The article concludes with recommendations for future directions.

## II. METHODOLOGY

This review follows established systematic review protocols [22]–[24], coupled with expert consultation in computer vision and driver behavior [25]. The methodology comprises key steps, including study selection, inclusion and exclusion criteria, and final screening to ensure relevance of reviewed papers.

*Study Selection*

Please A comprehensive search was conducted across multiple academic databases, including IEEE Xplore, Google Scholar, ASCE Library, and TRB. The search was limited to peer-reviewed journal articles published in the last five years (2019-2024), with an emphasis on the most recent advancements in ML and DL applied to distracted driving detection. This period was chosen to capture the rapidly evolving nature of ML and DL techniques while acknowledging seminal works that have laid the foundation for current research. Seminal studies predating this range were included selectively if they contributed significantly to the conceptual development of the field.

The literature search employed a combination of keywords such as "distracted driving detection," "machine learning," "deep learning," "multimodal data," "naturalistic driving data," and "simulation-based studies."

*Inclusion and Exclusion Criteria*

The study ensured that only papers relevant to the study's objectives were included by introducing the following inclusion and exclusion criteria:

*Inclusion Criteria:* The review included studies that applied ML or DL techniques to distracted driving detection. Both public and private datasets were considered to provide comprehensive coverage. Studies incorporating distinct data modalities, such as visual, auditory, sensory, or multimodal approaches, were prioritized. Novel modalities offering unique perspectives on distraction detection were also considered.

*Exclusion Criteria:* Studies focused solely on crash prevention strategies without any ML or DL application were not included in this review. Articles that were not available in full text and in English were also omitted from the review.

*Screening Process*

The screening process involved a multi-step review of the identified articles. First, duplicate entries were removed. Next, titles and abstracts were reviewed to confirm relevance, followed by a full-text review for more in-depth assessment. The final selection process ensured that only articles directly aligned with the objectives of this review—ML/DL techniques applied to different modalities for distracted driving detection—were included.

*Total Number of Papers:* The initial search across various databases yielded a total of 606 articles. After a thorough application of screening and inclusion criteria, which focused on relevance and the utilization of machine learning (ML) or deep learning (DL) techniques, 285 articles were excluded. This left 321 articles to undergo a full-text review, resulting in 75 papers being selected for inclusion in this review. It is important to note that while 75 papers were included in the final review, only 54 of these were specifically categorized according to the focus of the study, with the remaining 21 papers being referenced more broadly in other sections of the paper.

Figure 1 provides a detailed review of the selection and inclusion process utilizing the Preferred Reporting Items for Systematic Reviews and Meta-Analyses (PRISMA) framework provided Haddaway et al. [26].

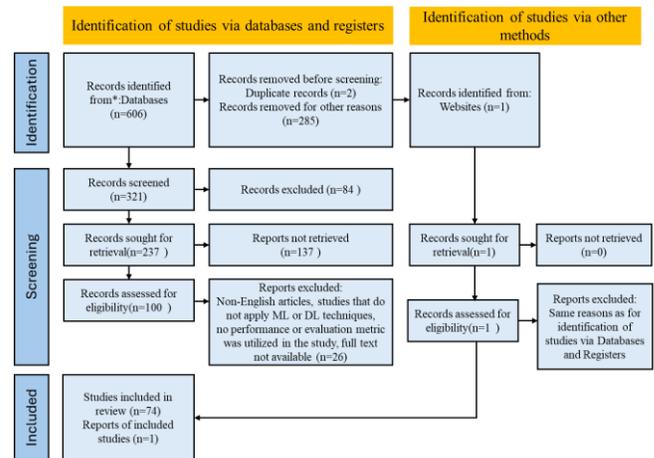

Fig. 1. PRISMA Flow Diagram: Study Identification and Screening Process.

## III. CATEGORIZATION OF THE REVIEWED PAPERS

In the review of the extant literature, studies employed the use of publicly available datasets or private datasets. The reviewed papers were therefore grouped based on whether they used publicly available datasets or private datasets. Figure 2 provides some common examples of public and private datasets used extensively in distracted driving studies.

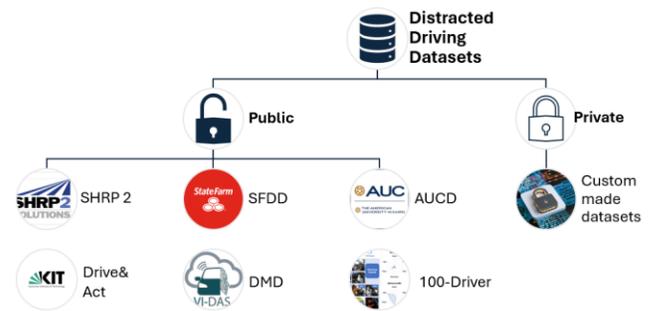

Fig. 2. Public and Private Datasets

Under each group of datasets utilized for the study, papers were further categorized based on the data modality employed in distracted driving detection including visual, auditory, sensory(signal), or multimodal (combining two or more of the distinct modalities). Figure 3. presents the breakdown of papers by the type of dataset used (public or private), data modality, and the number of papers that applied ML/DL techniques in each category.

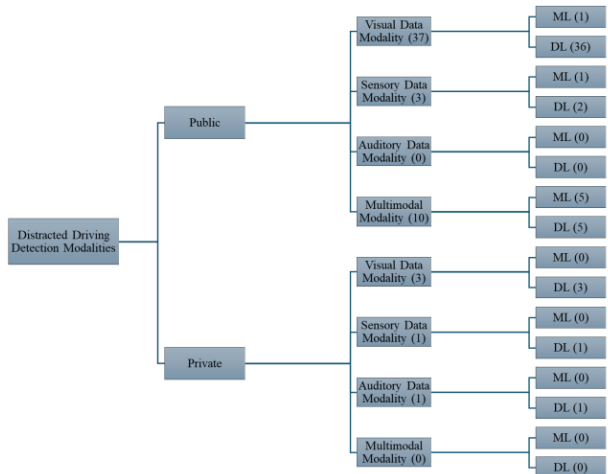

Fig. 3. Categorization of the Reviewed Papers

This categorization provides the basis for assessing the strengths and weaknesses of the methods employed for each modality, providing a comparative view of their contribution to detecting distracted driving. By organizing the studies in this way, we can assess how different data modalities and techniques perform in analyzing distracted driving behavior.

For each data modality, the *ML or DL techniques* employed are further discussed in subsequent sections to evaluate their effectiveness in detecting distracted driving behaviors. This approach ensures that readers gain a clear understanding of which modalities and techniques offer the most comprehensive insights into driver's behavior regarding distraction during real-world driving conditions.

### Public Datasets

Various agencies and organizations have curated and publicly released datasets to support research efforts in detecting distracted driving. These datasets are essential for benchmarking and training machine learning models, offering open access to diverse, standardized data and enabling comparability across studies. Such efforts aim to enhance ADAS by alerting and assisting drivers in avoiding distractions on the road.

This section summarizes publicly available datasets used in distracted driving detection shown in Table 1. Following this, the modalities provided by publicly available datasets utilized by various research studies—*visual, sensory, auditory*, and *multimodal*—are discussed. The ML or DL techniques employed are examined to assess the effectiveness of these techniques across the different modalities, identifying which modality offers a more comprehensive detection of distracted drivers.

TABLE I
COMPARISON OF PUBLICLY AVAILABLE DATASETS FOR DISTRACTED DRIVING DETECTION AND CLASSIFICATION

| | SHRP 2 | StateFarm | AUC | EBDD | 3MDAD | Drive&Act | DMD | 100-Driver |
|---|---|---|---|---|---|---|---|---|
| **Year** | 2015 | 2016 | 2017 | 2018 | 2019 | 2019 | 2020 | 2022 |
| **# of Views** | 5 | 1 | 1 | 1 | 2 | 6 | 3 | 4 |
| **Camera Streams** | GRAY/IR | RGB | RGB | RGB | RGB, Depth | RGB, Depth, IR | RGB, Depth, IR | RGB, NIR |
| **Resolution** | 356×240 | 640×480 | 1920×1080 | 854×480 | 640×480 | 1920×1080 | 1920×1081 (RGB), 1280×720 (IR/Depth) | 1920×1080 |
| **# of Classes** | 6 | 10 | 10 | 4 | 16 | N/A | N/A | 22 |
| **Size** | 91K | 22K | 17K | N/A | N/A | N/A | N/A | 470K |
| **# of Participants** | over 3000 | 26 | 31 | 13 | 50 | 15 | 37 | 100 |
| **# Male/Female** | N/A | N/A | 22/9 | 13/0 | 38/12 | 11/4 | 27/10 | 70/30 |
| **Video Duration** | over 1 M h | N/A | N/A | ~ 0.67 h | ~ 6.12 h | ~ 12 h | ~ 41 h | ~ 79.34 h |
| **Modalities** | Multimodal (Visual, sensory, auditory) | Visual | Visual | Visual | Multimodal (Visual, sensory) | Multimodal (Visual, sensory) | Multimodal (Visual, Sensory) | Multimodal (Visual, Sensory) |

### Visual Data Modality

Visual data from in-vehicle cameras has emerged as the *most widely* used modality for distracted driving detection [3], capturing real-time cues such as *facial expressions, eye gaze, head posture, arm and body movements* [27]. These indicators enable detection of various distractions, including manual (e.g., phone use) and visual (e.g., eyes off the road) behaviors. The rich information captured through visual data has led researchers to explore diverse ML and DL approaches for effective analysis. Common techniques include CNNs for image processing, temporal models like *Long Short-Term Memory* (LSTM) networks for sequential analysis, and *hybrid architectures* combining multiple approaches for improved performance. These methods have demonstrated capability in processing complex visual information to detect and classify different types of distracted behaviors in both offline analysis and real-time applications. This section examines how these techniques have been applied to *visual data* for distracted driving detection, highlighting key methodological approaches and implementations. Figure 4 gives an overview of the various techniques captured in this review.

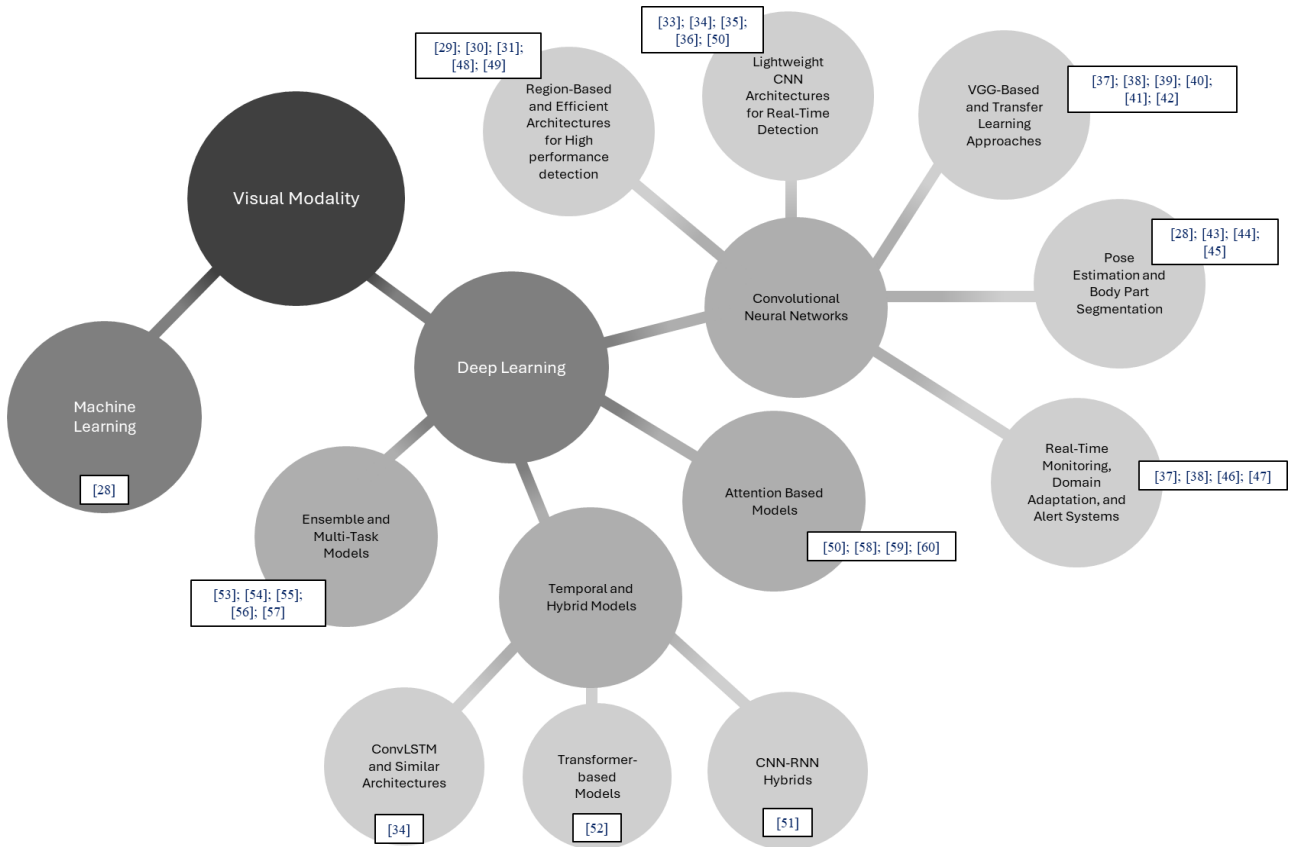

Fig. 4. Summary of Techniques Applied to Visual Data Modality

*Methods Used*

The classification and detection of distracted driving behaviors using visual data can be broadly grouped into ML and DL approaches.

Machine Learning (ML)

In recent years, few studies have employed ML techniques. For example, Gong and Shen [28] proposed an algorithm using pose features and an adjustable distance-weighted KNN (ADW-KNN) for distracted driving detection. The model extracts key points from driver poses using OpenPose and classifies behaviors with ADW-KNN. Tested on the State Farm Distracted Driver Detection (SFD3) dataset, it achieved 94.04% accuracy and 50 FPS, providing a fast, effective solution for real-time monitoring.

However, the approach presents some limitations. The algorithm requires *iterative parameter* tuning for *P* and *K* values, increasing computational complexity and potentially hindering practical deployment. The need for larger K values compared to traditional KNN further challenges the method's scalability, particularly for extensive datasets. These constraints highlight critical areas for future research in machine learning-based distracted driving detection.

Deep Learning (DL)

Deep learning has revolutionized computer vision and pattern recognition, making it an essential approach for complex tasks such as distracted driving detection. Visual data modalities, including video frames and images, benefit greatly from deep learning's ability to accurately identify and classify driver behavior, enhancing the robustness and reliability of detection systems. This section outlines the *various deep learning techniques* applied specifically to visual data modality for the detection of distracted driving. The approaches are categorized into four distinct but interconnected techniques: CNN-based approaches, temporal and hybrid models, ensemble and multi-task models, and attention-based models. Each subsection will provide detailed descriptions of the architectural frameworks used by a variety of studies, as well as the strengths and weaknesses associated with each approach.

*CNN-Based Approaches*

CNNs have emerged as the predominant deep learning architecture for driver distraction detection, demonstrating significant evolution from 2019 to 2024. The field has progressed along three main trajectories: *increasing detection accuracy* (from ~90% to >99% on standard benchmarks), *reducing computational complexity* (from large architectures like VGG to efficient designs like MobileNetV3), and *enhancing real-time performance* (achieving inference speeds up to 90 FPS). Recent trends show a shift from general-purpose CNNs to specialized architectures that balance accuracy with deployment constraints. These advances are characterized by: (1) *region-based approaches* that focus on critical areas like the driver's face and hands [29], [30], [31], [48], [33], (2) *lightweight architectures* optimized for edge devices and real-time detection [34]–[37], (3) *transfer learning* strategies that leverage pre-trained models [38], [39], [40], [41], [42], [43], (4) *pose estimation techniques* for precise body part tracking [44], [28], [45], [46], and (5) *domain-adaptive* methods that improve generalization across different driving environments [38], [39], [47], [48]. This progression reflects the field's movement toward more practical, deployable solutions for real-world driver monitoring systems. The following subsections present the findings of several studies that employ CNN-based approaches.

Region-Based and Efficient Architectures for High-Performance Detection.

Region-based CNN approaches, such as Faster R-CNN, have demonstrated significant potential in isolating *key regions* like the *driver's face, body*, and *hands* to detect distractions effectively. For instance, Wang et al. [29] improved detection accuracy from 95.31% to 96.97% by leveraging Faster R-CNN to focus on the driver's upper body and steering wheel while removing irrelevant background features. The classification of these extracted regions using Xception enhanced the model's ability to identify relevant visual cues. Despite its accuracy, the reliance on region proposal networks increased computational demands, limiting its real-time deployment feasibility and performance in cases of overlapping or subtle distractions.

In comparison, Sajid et al. [30] found EfficientDet-D3 to outperform Faster R-CNN and YOLO-V3 with a mean Average Precision (mAP) of 99.16%. EfficientDet's *hierarchical scaling* and *feature fusion* capabilities allowed it to effectively capture manual distractions, such as texting and operating a radio. However, its focus on visual distractions makes it less adept at detecting non-visual or cognitive distractions, which are equally critical for comprehensive distracted driving detection systems.

Similarly, Zheng et al. [49] introduced the CornerNet-Saccade model, which achieved 97.2% accuracy by focusing on critical regions, including the driver's face and hands. This approach proved particularly robust for detecting subtle distractions like eating or phone usage. However, its performance in uncontrolled real-world environments remains unproven, and the model requires further validation under varied lighting and occlusion conditions to ensure scalability.

Efficient architectures continue to push the boundaries of detection performance. Peruski et al. [31] demonstrated the superiority of EfficientNetB7, which achieved an AUC of 1.00 for front-seat passenger detection, surpassing Faster R-CNN in terms of precision. Despite its exceptional performance, EfficientNetB7 shares the limitation of being data-intensive, requiring high-quality, preprocessed datasets to maintain its accuracy. Additionally, its computational requirements, though optimized compared to traditional models, still present challenges for integration into real-time systems.

Finally, Fan and Shangbing [50] developed an optimally weighted multi-scale local feature fusion network for driver distraction recognition, combining global image features (extracted by ResNet-50) with localized features detected through YOLOv5. This fusion strategy achieved 94.34% accuracy on the HY Large Vehicle Driver Dataset and 95.84% on the AUC dataset, demonstrating robustness in differentiating visually similar distractions, such as texting versus talking on the phone. However, the complexity of the fusion process increases computational overhead, making real-time applications challenging without further optimization.

Figures 6 and 7 illustrate the generalized architectures for region-based and efficient detection approaches, respectively. These frameworks highlight the critical stages, from ROI extraction to classification, used across the reviewed studies.

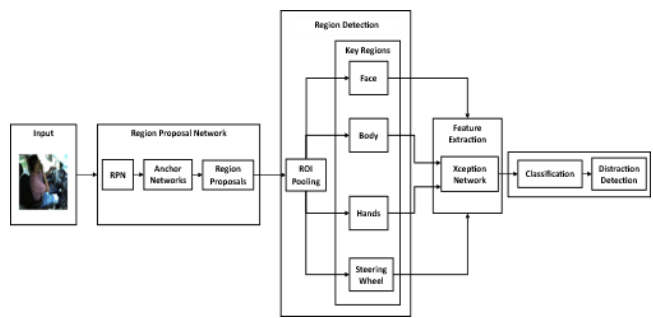

Fig. 6. General Architecture for Region-Based Detection. Adapted from Wang et al. [29].

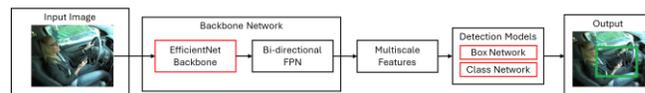

Fig. 7. General Architecture for EfficientNet. Adapted from Sajid et al.[30].

Table II presents a quick snapshot of the strengths and weaknesses for region-based and efficient architectures identified in selected studies.

| Strengths | Weaknesses |
|---|---|
| • High detection accuracy for key regions (e.g., face, hands, steering wheel) [29, 30, 48, 49] | • Limited to image-based data, reducing effectiveness for detecting non-visual distractions (e.g., cognitive) [30, 31] |
| • Enhanced feature extraction using hierarchical scaling and fusion (e.g., EfficientDet-D3) [30] | • Performance depends heavily on high-quality, preprocessed datasets [29, 31] |
| • Robust performance in distinguishing visually similar distractions (e.g., texting vs. talking) [49] | • Struggles with co-existing or overlapping distractions and subtle behavior variations [48, 49] |
| • Preprocessing optimizes detection accuracy by removing irrelevant distractions (e.g., Faster R-CNN) [29] | • Computational complexity hinders real-time deployment (e.g., Faster R-CNN, EfficientNetB7) [29, 30, 31] |
| • Demonstrates scalability in achieving state-of-the-art metrics (e.g., mAP, AUC) [30, 31] | • Requires further validation in real-world, uncontrolled environments (e.g., lighting, occlusion) [48, 49] |

### Lightweight CNN Architectures for Real-Time Detection

The development of lightweight CNN architectures for detecting distracted driving showcases a range of innovative solutions, each addressing real-time driver monitoring challenges while navigating specific limitations.

*SqueezeNet,* presented by Sahoo et al. [34] , stands out as a compact and efficient model, achieving an impressive 99.93% accuracy on the SFDDD dataset. Its small size makes it ideal for deployment on resource-limited devices like the Raspberry Pi 4B. However, its evaluation is limited to *static conditions*, leaving its performance in dynamic real-world driving scenarios unexplored.

Building on the goal of efficiency, Liu et al. [35] proposed a knowledge-distillation-based approach, reducing model size to just *0.42 million parameters* while maintaining a high accuracy of 99.86%. They also suggest expanding this work into a 3D CNN for analyzing driver behaviors over time, though this remains a conceptual promise yet to be realized.

López and Arias-Aguilar [36] introduced a TensorFlow Lite-based CNN focused on balancing accuracy and computational efficiency, achieving 91% accuracy on devices with limited processing power. However, *reducing computational demands comes at the cost of capturing intricate features*. Additionally, the dataset used primarily covers normal weather conditions, limiting the model's adaptability to adverse scenarios like rain or fog.

*The D-HCNN model,* developed by Qin et al. [37] , uses Histogram of Oriented Gradients (HOG) preprocessing to filter out background noise, enabling the model to concentrate on driver actions. This approach yields competitive results, with accuracies of 95.59% and 99.87% on different datasets. However, the increasing complexity of deeper convolutional layers poses challenges, such as filter sizes surpassing feature map dimensions, potentially limiting scalability.

Finally, *YOLOv8n,* proposed by Du et al. [51] , introduces advanced components like *GhostConv* and *GhostC2f layers*, reducing computational demands by 36.7%. This model also expands its dataset to include 14 distraction categories and nighttime conditions, enhancing versatility. However, these architectural innovations increase system complexity, potentially complicating integration with existing driver monitoring frameworks.

Each model offers unique strengths, from computational efficiency to high accuracy, yet all face challenges in balancing these goals with adaptability to real-world conditions. Together, they reflect the ongoing pursuit of a truly robust and reliable solution for detecting driver distractions across diverse scenarios. Figures 8,9, and 10 are sample flowcharts that depict lightweight architectures.

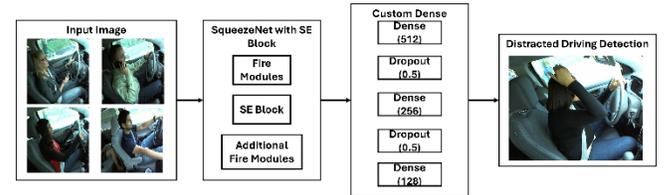

Fig. 8. SqueezeNet-Based Architecture with SE Block for Distracted Driving Detection. Adapted from Sahoo et al. [34].

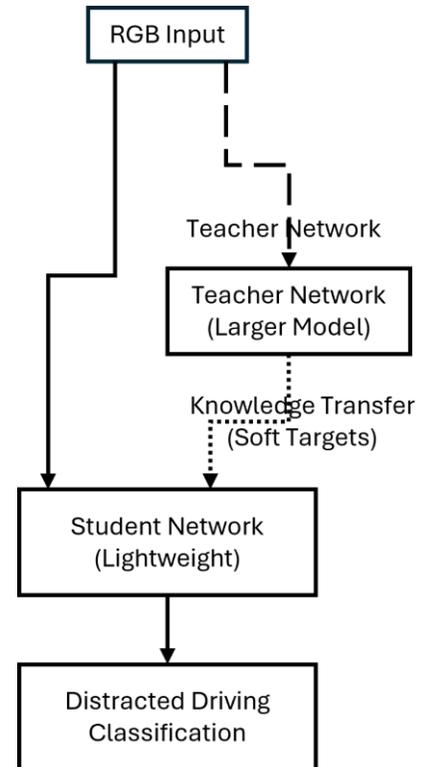

Fig. 9. Knowledge Distillation Framework for Lightweight Distracted Driving Detection Model. Adapted from Liu et al. [35].

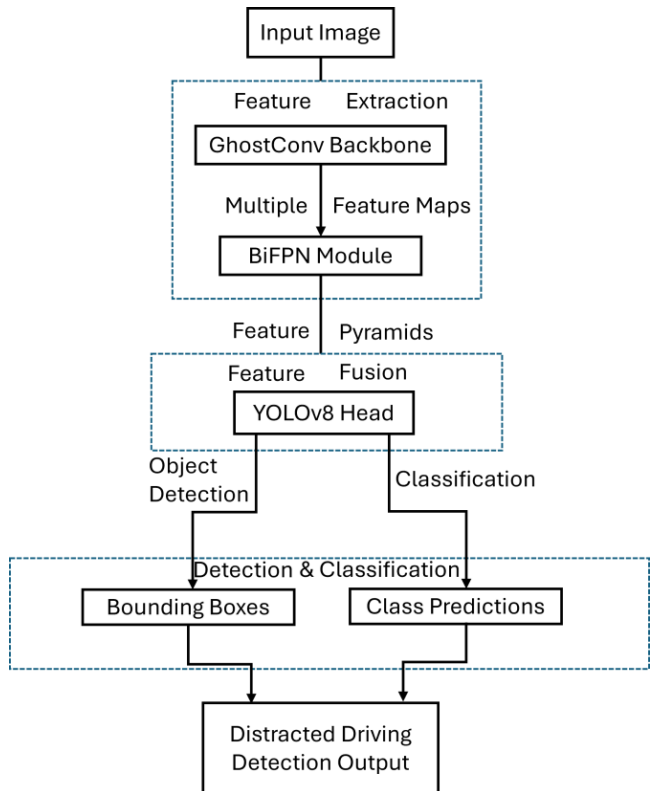

Fig. 10. Simplified Architecture of the Lightweight YOLOv8-Based Model for Distracted Driving Detection. Adapted from Du et al. [51].

Table III summarizes the strengths and weaknesses identified in the reviewed studies, providing an overview of their key contributions and limitations.

TABLE III
SRENGTHS AND WEAKNESSES OF LIGHTWEIGHT ARCHITECTURES

| Strengths | Weaknesses |
|---|---|
| • High accuracy with small model sizes (e.g., SqueezeNet, 99.93%) [33]. | • Limited testing in dynamic real-world scenarios [33]. |
| • Low parameter counts (e.g., 0.42 million in knowledge-distillation models) [34]. | • Scalability issues due to convolutional layer complexity (e.g., D-HCNN) [36]. |
| • Effective preprocessing techniques like HOG for noise reduction [36]. | • Limited performance in adverse weather conditions (e.g., TensorFlow Lite) [35]. |
| • Adaptable to nighttime scenarios and diverse distractions (e.g., YOLOv8n) [50]. | • Proposed 3D CNNs for temporal analysis remain untested [34]. |
| • Optimized for low-power devices (e.g., TensorFlow Lite CNNs) [35]. | • Increased architectural complexity in models like YOLOv8n [50]. |

VGG-Based and Transfer Learning Approaches

The application of VGG-based models and transfer learning techniques has advanced the field of distracted driving detection with researchers like Anand et al. [38], Supraja et al. [39], Zhang and Ke [41], Oliveira and Farias [40], Goel et al. [42], and Vaegae et al. [43] making notable contributions to this critical area of technological innovation. These studies not only advance the technical capabilities of detection models but also lay the groundwork for practical, real-world applications that aim to enhance road safety through intelligent monitoring systems.

Anand et al. [38] and Supraja et al. [39] both utilized the VGG16 architecture, achieving impressive accuracy rates of over 90% and 93.75% respectively. Their approaches demonstrated promising real-time detection capabilities through alert mechanisms, yet simultaneously exposed fundamental limitations in dataset diversity and behavioral classification. The studies particularly *struggled* with *distinguishing* between *similar driving postures*, such as differentiating between safe driving and passenger interactions, highlighting the need for more diverse datasets and temporal data to enhance class differentiation.

Zhang and Ke [41] pushed the boundaries of model performance, incorporating *advanced regularization techniques* and *data augmentation* to achieve an exceptional 98.5% accuracy. Their work notably addressed overfitting challenges observed in previous architectures like ResNet50, while simultaneously highlighting persistent issues of dataset limitations and environmental variability. This work underscores the importance of tailoring models to mitigate biases in datasets and adapt to varying real-world conditions.

Oliveira and Farias [40] conducted a comprehensive comparative analysis across multiple architectures, including VGG19, DenseNet161, and InceptionV3. Their research critically revealed the superiority of end-to-end optimization, with DenseNet161 achieving 88.83% accuracy. More importantly, their study demonstrated that alternative transfer learning strategies—such as *limiting optimization to fully connected layers* or *using convolutional layers merely as feature extractors*—resulted in significantly reduced generalization, with test accuracies plummeting to 57.41% and below 30%, respectively. These findings not only highlight the need for optimizing pre-trained layers but also suggest that simplifying architectures without considering task-specific nuances can compromise performance.

Goel et al. [42] explored lightweight architectures, finding MobileNetV2 particularly promising with a 93.8% accuracy rate. Their approach distinguished itself by developing a user-interactive web application, showcasing the potential for integrating lightweight and efficient models into practical use cases. However, like many other studies, their research was constrained by *reliance on limited datasets* and a narrow focus on visual distractions. Expanding beyond visual data to include cognitive and auditory modalities could enhance the comprehensiveness of such systems, enabling them to address a broader spectrum of distracted driving behaviors.

Vaegae et al. [43] provided a nuanced comparison between VGG-16 and ResNet-50 architectures, achieving accuracies of 86.1% and 87.92% respectively. Their research critically examined the challenges of overfitting, dataset bias, and the computational intensiveness of neural network architectures in edge-based vehicle systems. By addressing these challenges, they underscored the importance of integrating multimodal inputs, including physiological and environmental data, to create robust and adaptable distracted driving detection systems. This shift toward multimodal integration marks a step forward in addressing non-visual distraction modalities that remain largely overlooked in many existing models.

A consistent narrative emerges across these studies: while VGG-based models and transfer learning techniques demonstrate remarkable potential in distracted driving detection, significant methodological challenges persist. These include *dataset constraints*, *difficulties in capturing temporal* and *multimodal data*, and *the exclusion of non-visual distraction modalities*. And the collective trajectory of this research reveals an evolution from purely technical exploration to practical deployment considerations, such as Goel et al.'s web application and Vaegae et al.'s focus on edge-based implementation.

These studies collectively highlight the potential for advancing distracted driving detection while also underscoring the need for addressing current limitations, which will be explored further in the conclusion. Figures 11,12,13, 14 and 15 are selected methodological approaches employed by scholars of reviewed studies in this section.

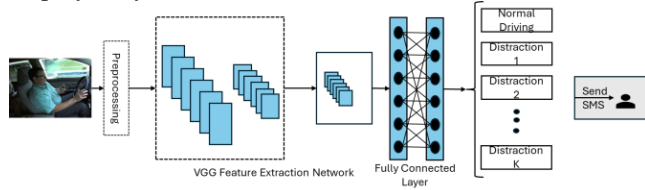

Fig. 11. VGG-Based Visual Feature Extraction and Classification Framework for Distracted Driving Detection. Adapted from Supraja et al. [39].

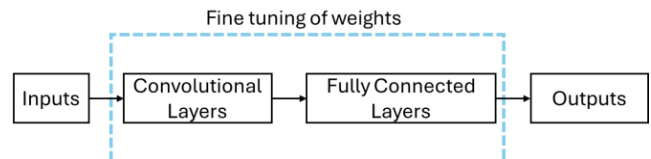

Fig. 12. Transfer learning with end-to-end fine-tuning approach. Adapted from Oliveira and Farias [40].

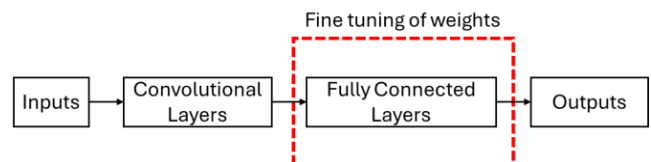

Fig. 13. Transfer learning with fully connect layers fine tuning. Adapted from Oliveira and Farias [40].

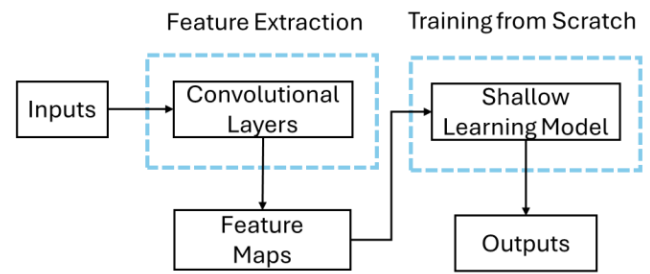

Fig. 14. Transfer learning by feature extraction approach. Adapted from Oliveira and Farias [40].

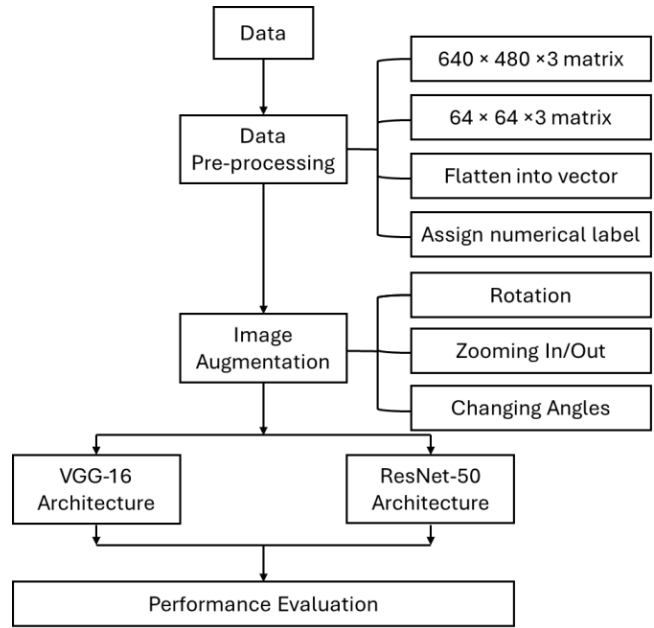

Fig. 15. Workflow for Distracted Driver Detection Using Transfer Learning with VGG-16 and ResNet-50 Architectures. Adapted from Vaegae et al. [43].

Table IV summarizes the strengths and weaknesses identified in the reviewed studies of VGG-based and transfer learning approaches to visual data modalities.

| Strengths | Weaknesses |
|---|---|
| • Achieved high accuracy across multiple architectures (e.g., VGG16: 90%–98.5%, DenseNet161: 88.83%) [37, 39, 40]. | • Limited dataset diversity reduces generalizability to real-world scenarios with varied distraction behaviors [37, 38, 40]. |
| • Effective use of transfer learning and end-to-end optimization techniques for robust feature extraction [39, 41]. | • Struggles with distinguishing visually similar behaviors, such as safe driving versus passenger interaction [37]. |
| • Incorporates regularization and augmentation techniques to address overfitting in models (e.g., L2, data flips) [40]. | • Lack of temporal data integration limits the ability to capture sequential distraction patterns [37, 41, 42]. |
| • Architectures like MobileNetV2 demonstrated feasibility for practical applications [41]. | • Heavy reliance on visual distractions excludes cognitive and auditory modalities, reducing system comprehensiveness [41, 42]. |
| • Practical implementations such as web applications showcase readiness for deployment [41]. | • Computationally intensive models (e.g., ResNet50) hinder deployment in edge-based vehicle systems [42]. |
| • Pre-processing techniques (e.g., resizing, augmentation) improve model training and reduce dataset bias [42]. | • Environmental factors like lighting variability remain largely unaddressed, affecting real-world robustness [38, 40]. |

**Pose Estimation and Body Part Segmentation**

The integration of *pose estimation* and *body part* segmentation techniques has significantly advanced driver distraction detection by enabling systems to focus on critical regions, such as the *head, hands,* and *posture*. These approaches leverage deep learning and traditional machine learning methodologies to enhance detection accuracy and efficiency, addressing a key challenge in isolating distractions effectively.

Several studies have demonstrated the utility of segmenting driver body parts for targeted analysis. Ezzouhri et al. [44] employed deep learning-based segmentation to extract features from the driver's head and hands, achieving 96.25% accuracy on an internal dataset and over 95% on the AUC benchmark using VGG-19 and Inception-V3 classifiers. This segmentation significantly improved classification performance by reducing background noise and narrowing focus on relevant areas. Similarly, Du et al. [45] utilized YOLOv5-GBC with GhostConv and BiFPN to segment key regions, achieving a 91.8% mAP on an extended dataset. Both studies emphasized the value of segmentation in improving model precision, although dataset limitations, such as a lack of diversity in environmental conditions and driver characteristics, constrained their generalizability.

Pose estimation techniques further advanced this field by enabling real-time detection capabilities. Gong and Shen [28] employed OpenPose integrated with an enhanced ADW-KNN classifier, achieving a high accuracy of 94.04% at 50 FPS. This approach balanced computational efficiency and accuracy, making it suitable for real-time applications. However, the reliance on *parameter tuning* for ADW-KNN introduced computational complexity, and the dataset's limited perspectives hindered the model's ability to generalize across different viewing angles. In contrast, Deepthi et al. [46] focused on facial landmarks such as the Eye Aspect Ratio (EAR) and Mouth Aspect Ratio (MAR) to detect driver drowsiness, emphasizing the real-time practicality of their system. Their CNN-based approach performed robustly under varied lighting conditions, providing real-time alerts, but similar to other studies, faced challenges with dataset diversity and adaptability to real-world scenarios.

A recurring theme among these studies is the critical role of *data diversity* and *multimodal integration*. While segmentation and pose estimation techniques effectively enhance model focus and reduce computational overhead, their reliance on narrowly scoped datasets limits their robustness. For instance, both Ezzouhri et al. [44] and Deepthi et al. [46] highlighted the importance of incorporating diverse driver demographics, clothing variations, and environmental conditions (e.g., lighting, weather) to improve generalizability. Du et al. [45] similarly proposed enriching datasets with nighttime and adverse weather scenarios to address the limitations of their daytime-focused data.

In terms of methodological innovation, the use of multimodal data emerges as a promising direction for enhancing detection systems. Deepthi et al. [46] suggested integrating inputs from steering wheel sensors and lane departure warnings, while Du et al. [45] advocated for combining radar and image data to provide richer information. These multimodal approaches could mitigate the limitations of relying solely on visual cues, as demonstrated by Gong and Shen's [28] ADW-KNN algorithm, which struggled with computational demands despite its robust performance.

A key area of convergence across these studies is the move toward lightweight architectures and edge-based deployment. Gong and Shen [28] and Deepthi et al. [46] emphasized the need for deploying systems on edge devices to reduce reliance on cloud infrastructure and enhance real-time performance. Similarly, Du et al. [45] proposed optimizing lightweight models and incorporating attention mechanisms to improve computational efficiency without sacrificing detection accuracy. These efforts align with the broader goal of making driver monitoring systems practical for real-world applications, particularly in vehicles with limited computational resources.

In summary, pose estimation and body part segmentation have proven to be valuable tools for driver distraction detection, enhancing accuracy by narrowing focus to relevant features. However, challenges related to dataset diversity, real-world adaptability, and multimodal integration persist. Future research must address these limitations by incorporating more comprehensive datasets, leveraging multimodal inputs, and optimizing models for deployment on edge devices. By bridging the gap between technical innovation and practical application, these advancements can pave the way for more reliable and robust distracted driving detection systems. Figures 16 and 17 are selected examples of architectures under this category.

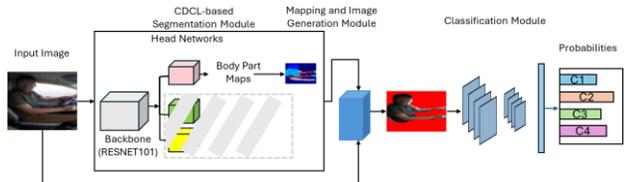

Fig. 16. Pose Estimation and Body Parts Segmentation Workflow. Adapted from Ezzouhri et al. [44].

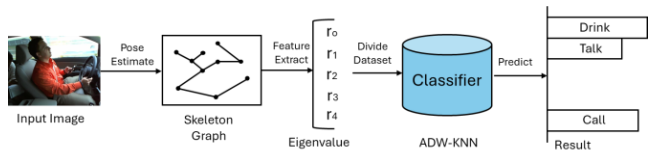

Fig. 17. Pose Estimation and ADW-KNN Classifier Workflow for Driver Distraction Detection. Adapted from Ezzouhri et al. [28].

Table V summarizes the strengths and weaknesses identified in the reviewed studies of pose estimation and body part segmentation approaches to visual data modalities.

TABLE V
STRENGTHS AND WEAKNESSES OF POSE ESTIMATION AND BODY PART SEGMENTATION

| Strengths | Weaknesses |
|---|---|
| • Improved focus on relevant body parts like the head and hands, enhancing detection accuracy [43, 44]. | • Datasets lack diversity, particularly in terms of environmental conditions, driver demographics, and clothing [43, 45]. |
| • Effective integration of pose estimation techniques (e.g., OpenPose) for real-time detection [28]. | • Limited generalizability due to reliance on datasets with fixed perspectives and daytime conditions [28, 44]. |
| • High performance with lightweight models suitable for edge-based deployment [28, 45]. | • Parameter tuning in algorithms like ADW-KNN increases computational complexity [28]. |
| • Robust performance under varied lighting conditions (e.g., facial landmarks for drowsiness detection) [45]. | • Absence of multimodal integration limits the ability to detect cognitive or non-visual distractions [45]. |
| • Innovative use of segmentation to reduce noise and focus on key features (e.g., YOLOv5-GBC with GhostConv and BiFPN) [44]. | • Lack of nighttime and adverse weather scenario data limits real-world applicability [44]. |
| • Real-time alert capabilities enhance practical application in distraction detection [45]. | • Segmentation models like VGG-19 rely heavily on computational resources, affecting scalability [43]. |

**Real-Time Monitoring, Domain Adaptation, and Alert Systems**

Real-time monitoring systems with integrated alert mechanisms play a pivotal role in enhancing the practicality of driver distraction detection. Lin and Hsu [47] developed a real-time monitoring system leveraging MobileNetV3 and a modified Keypoint Feature Pyramid Network (KFPN), achieving 95.6% accuracy and operating at an impressive 90 FPS. This high frame rate ensures the system's suitability for real-time applications, addressing the critical need for prompt detection and alerts. However, the system exhibited several limitations. It did not consider environmental factors such as weather conditions, road surfaces, or variations in speed limits, all of which are crucial for driving safety. Additionally, the framework's simplicity, while easy to implement, limited its ability to account for the behavior of surrounding vehicles, such as sudden braking, or more nuanced distractions like caring for children. The system also relied on the State Farm dataset, which includes only 10 distraction categories, restricting its adaptability to real-world scenarios involving diverse driver behaviors. Furthermore, the system's real-time detection performance was dependent on maintaining vehicle speeds below 81 km/h, limiting its utility in high-speed driving environments.

To address generalization issues across datasets, Wang and Wu [48] introduced a multi-scale domain adaptation network (MSDAN), achieving 96.82% accuracy on the State Farm dataset and 94.30% accuracy on the AUCD dataset, outperforming standard CNN models by up to 6.53%. The MSDAN approach incorporates a backbone network for enhanced feature extraction, a domain adaptation network for handling data source differences, and dropout methods to improve generalization. However, the study identified limitations in the recognition of distractions across datasets with the same distraction class, camera angle, and consistent data modality. This reveals the challenge of poor cross-dataset generalization, which remains a critical bottleneck in the robustness of distraction detection systems. Additionally, the limited data volume in existing open-source datasets constrains the model's potential to generalize effectively. The authors proposed future directions, including constructing large-scale, multi-angle, and multi-label datasets and extending their approach to capture spatiotemporal features through video and real-time camera data.

The real-time components of previously discussed systems, such as Anand et al. [38] and Supraja et al. [39], further emphasize the value of integrating alert mechanisms into monitoring frameworks. Anand et al. incorporated SMS-based notifications to promptly warn drivers of distracted behavior, while Supraja et al. designed a practical system for sending alerts to drivers and corresponding authorities, demonstrating the feasibility of real-time intervention. Figure 18 provides a classic example of a real-time monitoring workflow.

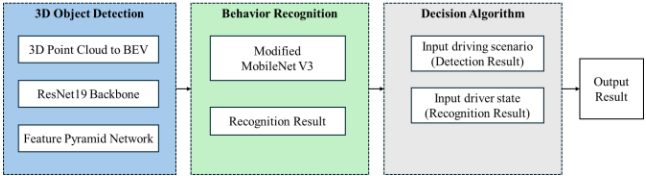

Fig. 18. Real Time Monitoring Workflow for Driver Distraction Detection. Image Adapted from Lin and Hsu [47].

Table VI summarizes the strengths and weaknesses identified in the reviewed studies of real-time monitoring, domain adaptation, and alert systems approaches to visual data modalities.

TABLE VI
STRENGTHS AND WEAKNESSES OF REAL-TIME MONITORING, DOMAIN ADAPTATION, AND ALERT SYSTEMS

| Strengths | Weaknesses |
|---|---|
| • High accuracy rates achieved with advanced models like MSDAN (96.82% on State Farm, 94.30% on AUCD) [47]. | • Limited generalization for cross-dataset scenarios, particularly for similar distraction classes [47]. |
| • Real-time detection capabilities demonstrated with MobileNetV3 operating at 90 FPS [46]. | • Lack of consideration for environmental factors, such as weather, road conditions, and speed limits [46]. |
| • Integration of domain adaptation techniques improves robustness across datasets [47]. | • Limited datasets (e.g., State Farm, with only 10 distraction categories) restrict model adaptability [46, 47]. |
| • Practical implementation of alert systems via SMS and real-time notifications [37, 38]. | • System dependencies on specific vehicle speeds (below 81 km/h) affect real-time detection in high-speed scenarios [46]. |
| • Multi-scale domain adaptation improves feature extraction and dataset adaptability [47]. | • Simplistic frameworks fail to incorporate interactions with surrounding vehicles or complex driver behaviors [46]. |
| • Dropout methods enhance model generalization performance in domain adaptation networks [47]. | • Real-time performance is computationally intensive, limiting scalability on embedded systems [46, 47]. |

*Temporal and Hybrid Models*

Detecting driver distraction not only involves analyzing static images but also requires understanding *temporal dependencies* and *dynamic behaviors.* Temporal and hybrid models have emerged to capture these dependencies, combining spatial-temporal features using deep learning techniques such as RNNs, LSTMs, ConvLSTMs, and advanced frameworks like Transformers. The subsequent subsections under temporal and hybrid models will discuss studies that have utilized these approaches.

*ConvLSTM and Similar Architectures*

Liu et al. [35] extended their lightweight student network into a spatial-temporal 3D Convolutional Neural Network (CNN) for video-based driver behavior recognition, achieving notable performance improvements on the Drive&Act dataset with only 2.03 million parameters. By leveraging knowledge distillation and neural architecture search (NAS), the model effectively captured spatial-temporal features while maintaining computational efficiency. However, its reliance on pre-existing datasets and the need for precise 3D convolutional kernel settings limited its generalizability to diverse driving environments. Additionally, the computational demands of video data training remain a challenge for broader real-time applications.

*CNN-RNN Hybrids*

Kumar et al. [52] proposed a genetically optimized ensemble of six deep learning models, combining AlexNet, VGG-16, EfficientNet B0, Vanilla CNN, Modified DenseNet-201, and a hybrid InceptionV3 + BiLSTM. By leveraging a genetic algorithm to assign weights to each model's output, the ensemble effectively balanced spatial feature extraction from CNN models and temporal dependency capture through the BiLSTM component. Tested on the AUC and State Farm datasets, the ensemble achieved 96.37% and 99.75% accuracy, respectively, demonstrating its effectiveness in recognizing subtle distractions such as phone usage or reaching behind. However, limitations include the need for deployment on embedded devices for real-time evaluation and addressing class confusion between similar behaviors, such as "Hair and Makeup" and "Talking to Passenger." Additionally, the ensemble could be utilized for generating labels for unlabeled images in the State Farm dataset, which would support more detailed studies. Figure 19 shows an example of a hybrid model.

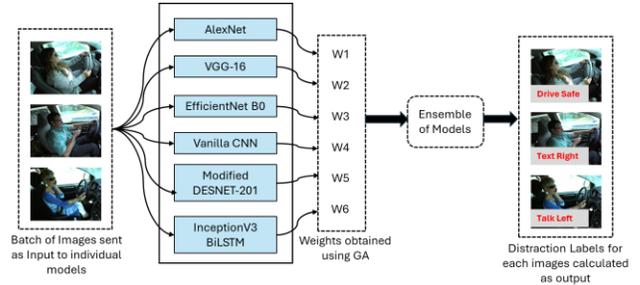

Fig. 19. Framework incorporating CNN-RNN Hybrid Architectures for Driver Distraction Detection. Adapted from Kumar et al. [52].

*Transformer-based Models*

Shi [53] introduced the DKT framework, which combines DWPose keypoint detection with a multi-transformer network to analyze driver behaviors. The framework leverages a High- and Low-Frequency Multi-Transformer Attention (HLMSA) mechanism to process temporal relationships between body key points at different frequencies, effectively capturing subtle distraction cues. By incorporating Kalman filtering, the model suppresses jitter in extracted keypoints, enhancing both spatial and temporal estimation. The DKT framework achieved 98.03% accuracy on the 100-Driver Dataset and demonstrated cross-dataset generalization with 73.88% accuracy on SFD2, outperforming several state-of-the-art models. Despite these advancements, the HLMSA mechanism's impact on accuracy was limited, with its primary contribution being a reduction in computational complexity. Furthermore, while DKT minimizes background noise influence and operates with only 1.76M FLOPs, its reliance on pre-trained keypoint models like DWPose and structured keypoint data may limit its adaptability to more complex or noisy environments. Future work could focus on enhancing generalization across datasets and exploring improvements in the accuracy of low-frequency distraction detection. Figure 20 provides the structural framework of DKT proposed in Shi's study.

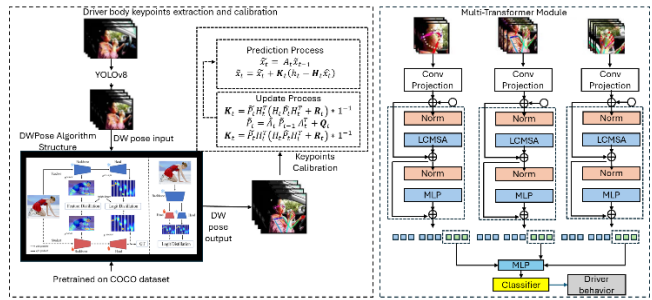

Fig. 20. The structural Framework of DKT. Image adapted from [53].

Table VII summarizes the strengths and weaknesses identified in the reviewed studies of temporal and hybrid models to visual data modalities.

<div align="center">

TABLE VII
STRENGTHS AND WEAKNESSES OF TEMPORAL AND HYBRID MODELS

</div>

| Strengths | Weaknesses |
|---|---|
| • Effective spatial-temporal feature extraction using 3D CNNs and hybrid CNN-RNN models [34, 51]. | • Computationally intensive training for video-based models limits real-time deployment [34]. |
| • Lightweight architectures (e.g., 3D CNN with 2.03M parameters, DKT with 1.76M FLOPs) enable efficiency [34, 52]. | • Dependency on pre-existing datasets limits generalizability to diverse or noisy environments [34, 52]. |
| • Genetic algorithm optimization balances model outputs for superior performance (96.37% on AUC, 99.75% on State Farm) [51]. | • Confusion between similar behavior classes (e.g., "Hair and Makeup" vs. "Talking to Passenger") reduces accuracy [51]. |
| • Transformer-based frameworks like DKT leverage advanced attention mechanisms for subtle distraction cues [52]. | • HLMSA's contribution to accuracy is minimal, emphasizing reduced computational complexity over direct accuracy gains [52]. |
| • CNN-RNN hybrids effectively capture temporal dependencies, making them suitable for real-time distraction detection [51]. | • Ensemble models require evaluation on embedded systems for practical real-time applications [51]. |
| • Cross-dataset generalization demonstrated by Transformer-based models (73.88% on SFD2) [52]. | • Reliance on structured keypoint data from pre-trained models (e.g., DWPose) affects adaptability to complex scenarios [52]. |

*Ensemble and Multi-Task Models*

Some advances in CNN architectures have shown a trend toward ensemble and hybrid approaches for enhanced distraction detection. These approaches can be categorized into three main strategies: hybrid CNN frameworks, weighted ensemble methods, and multi-task learning architectures.

In hybrid architectures, Huang et al. [54] demonstrated the effectiveness of combining ResNet50, Inception V3, and Xception through cooperative transfer learning, achieving 96.74% accuracy on the State Farm Dataset. This hybrid approach proved particularly effective at capturing subtle hand movements critical for distraction detection. However, the dataset used in this study only included images captured from the right-hand side, limiting the generalizability of the proposed Hybrid CNN Framework (HCF) when cameras are placed in different vehicle positions. Additionally, the framework's performance was significantly influenced by lighting conditions, reducing its accuracy in nighttime scenarios.

Building on this concept, Mollah et al. [55] developed a weighted SoftMax averaging ensemble of DenseNet121 and MobileNet, reaching 99.81% accuracy on the same dataset. The success of this approach highlighted how DenseNet's dense connectivity patterns complemented MobileNet's efficient depth-wise convolutions. However, while the model was economical with a combined 10.2 million parameters, the study noted the need to address overfitting through augmentation and proposed reducing parameters further by modifying the CNN models. These limitations suggest potential challenges in adapting the model to more diverse datasets or constrained environments.

Further innovations in ensemble methods emerged with Aljasim and Kashef's [56] E2DR model, which strategically combined ResNet50's computational efficiency with VGG16's deep feature extraction through a stacking ensemble approach. While achieving a modest 92% accuracy, the E2DR model demonstrated strong real-time performance capabilities. However, the study highlighted the computational complexity involved in developing the E2DR models from scratch, posing challenges for deployment in time-critical applications. Additionally, the study proposed exploring more combinations in the stacking ensemble method, which was infeasible due to limited computational resources. Future extensions include integrating the model with systems like driver authentication or alarm mechanisms and applying it in autonomous vehicles for health-related monitoring.

Subbulakshmi et al. [57] extended this direction by creating a more comprehensive ensemble incorporating ResNet50, VGG16, and DenseNet121, achieving 98.92% accuracy on the State Farm dataset. While this ensemble achieved high accuracy, the study acknowledged the need for further research into advanced ensemble modeling approaches to optimize performance. Future work could also focus on advanced data augmentation methods, transfer learning paradigms, and domain adaptation methodologies to enhance the ensemble model's applicability and generalizability across diverse datasets.

A notable departure from traditional ensemble approaches came from Liu et al.'s [58] Triple-Wise Multi-Task Learning (TML) framework. By introducing triplet-based image sampling and a multi-task strategy, TML achieved 96.3% accuracy on the AUC dataset and demonstrated robust cross-dataset performance with 66.9% accuracy on the Drive and Act Dataset. While the TML framework reduced local feature bias and improved generalization, it required approximately 90 million parameters to operate effectively. Although this is an improvement over previous state-of-the-art models (160 million parameters), further parameter reduction is necessary for real-world applications. Additionally, the model's best accuracy was achieved by averaging the logits from raw input and positive samples, suggesting potential for optimization to rely solely on raw input for performance enhancement. Figures 21 and 22 display examples of ensemble and multi-task models respectively.

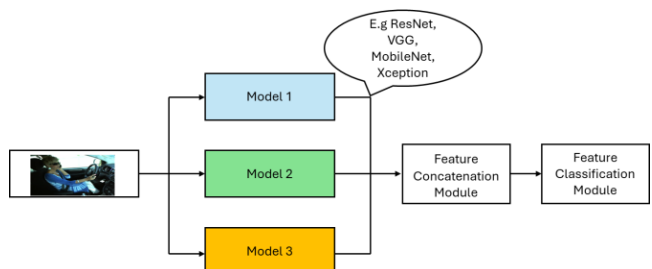

Fig. 21. Generic Ensemble Architecture for Distracted Driving Detection. Adapted from Huang et al. [53].

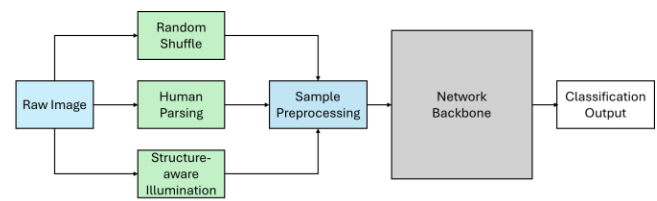

Fig. 22. Triple-Wise Multi-Task Learning Framework for Driver Distraction Detection. Adapted from [58].

Table VIII summarizes the strengths and weaknesses identified in the reviewed studies of ensemble and multi-task models to visual data modalities.

TABLE VIII
STRENGTHS AND WEAKNESSES OF ENSEMBLE AND MULTI-TASK MODELS

| Strengths | Weaknesses |
|---|---|
| • Hybrid architectures (e.g., ResNet50, Inception V3, Xception) effectively capture subtle hand movements [53]. | • Performance heavily influenced by lighting conditions, limiting accuracy at night [53]. |
| • Weighted ensemble methods like SoftMax averaging achieve high accuracy (99.81%) while balancing efficiency [54]. | • Overfitting remains a concern; requires advanced data augmentation to enhance generalization [54]. |
| • Stacking ensembles (e.g., E2DR) demonstrate strong real-time capabilities with modest accuracy (92%) [55]. | • Computational complexity of ensemble models poses challenges for real-time deployment [55]. |
| • Multi-task learning (e.g., TML) reduces local feature bias and enhances cross-dataset generalization [57]. | • High parameter count (90M) limits applicability for resource-constrained real-world applications [57]. |
| • Advanced ensemble combinations like DenseNet121 and MobileNet leverage complementary strengths for efficiency [54]. | • Models reliant on limited datasets (e.g., State Farm) struggle to generalize across diverse driving scenarios [53, 54]. |
| • Comprehensive ensembles like ResNet50, VGG16, and DenseNet121 achieve high accuracy (98.92%) [56]. | • Further optimization needed in ensemble modeling, domain adaptation, and transfer learning for scalability [56]. |
| • Novel multi-task frameworks achieve robust performance on cross-dataset tests (66.9% on Drive and Act) [57]. | • Optimal accuracy relies on averaging raw input and positive samples rather than raw input alone [57]. |

*Attention based Models*

Recent research has demonstrated significant advances in attention-enhanced architectures, particularly focusing on YOLO-based models and specialized attention mechanisms for driver monitoring. These developments reflect a crucial balance between detection accuracy and computational efficiency.

In YOLO-based architectures, Du et al. [51] made notable improvements to YOLOv8n through the integration of GhostConv and GhostC2f layers, optimizing computational efficiency while maintaining feature extraction capability. Their approach incorporated a bidirectional Feature Pyramid Network (FPN) and SimAM attention mechanism, specifically addressing the challenges of nighttime scenarios and expanding the traditional SFDDD dataset to 14 categories. However, the study's limitations include its reliance on datasets that primarily focus on daytime conditions, reducing its effectiveness in complex environments such as adverse weather or varying illumination conditions.

Building on YOLO architecture optimization, Li et al. [59] developed the AB-DLM model based on YOLOv5s, achieving 95.6% mAP at 71 FPS on the Driver Monitoring Dataset (DMD). The integration of Squeeze-and-Excitation (SE) attention modules with BiFPN demonstrated superior performance compared to lightweight models like YOLOv4-Tiny and PP-YOLO Tiny. However, the structural complexity of the model increased significantly due to the inclusion of SE and BiFPN modules, leading to a larger model with higher computational demands. Although this complexity did not compromise real-time detection, further training on more diverse datasets is necessary to improve the model's generalization.

Parallel advances in attention mechanisms have shown promising results. Ai et al. [60] introduced the Double Attention Convolutional Neural Network (DACNN), incorporating both spatial and channel attention to enhance feature discrimination. This dual-attention approach achieved 95.4% accuracy in detecting various distraction behaviors, demonstrating the effectiveness of targeted feature enhancement. However, the study focused heavily on maximizing immediate detection accuracy and computational efficiency, leaving questions about the model's adaptability in long-term or varied real-world scenarios. Future work emphasizes combining CNN with attention mechanisms for more robust distracted driving analysis.

Li et al. [61] further advanced attention-based approaches with SWAM (Shifted Window Attention Mechanism), achieving 93.97% accuracy on the AUC Distracted Driver Dataset V2. SWAM leveraged self-attention to focus on important regions and capture long-distance dependencies efficiently. However, the study acknowledged the limitations of CNN architectures, particularly their reliance on local receptive fields, which pose a trade-off between speed and accuracy. Additionally, further exploration of combining CNNs with self-attention mechanisms was suggested to address this inherent constraint.

These developments highlight a clear trend toward sophisticated attention mechanisms that can improve detection accuracy while maintaining real-time performance capabilities. The success of these models demonstrates the importance of balancing architectural complexity with computational efficiency in practical driver monitoring systems, while future work should address generalization challenges and scalability in diverse real-world conditions. Examples of architecture or frameworks for attention-based models are shown in Figures 23, 24, and 25.

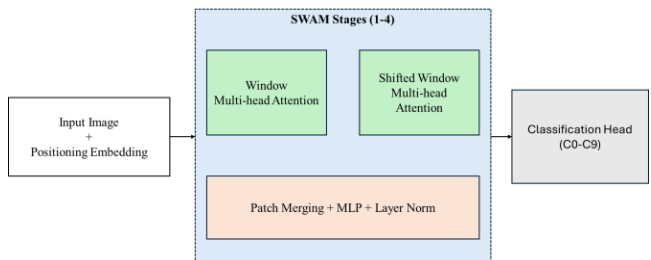

Fig. 23. An overview of SWAM architecture. Adapted from [61].

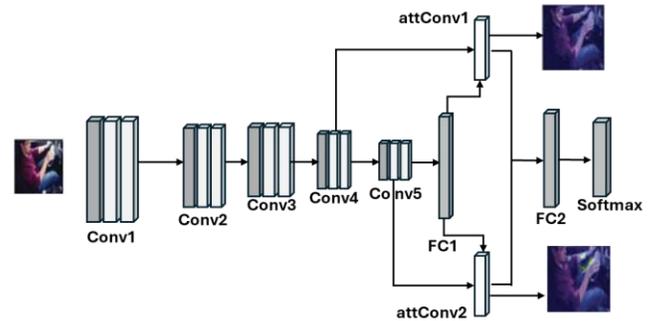

Fig. 24. Architecture of Double Attention CNN Model. Adapted from Ai et al. [60].

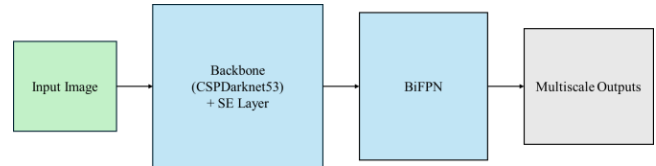

Fig. 25. The pipeline of AB-DLM network. Adapted from Li et al. [59].

Table IX summarizes the strengths and weaknesses identified in the reviewed studies of attention-based models to visual data modalities.

TABLE IX
STRENGTHS AND WEAKNESSES OF ATTENTION-BASED MODELS

| Strengths | Weaknesses |
|---|---|
| • Enhanced feature extraction using attention mechanisms like SimAM, SE, and BiFPN for improved accuracy [50, 58]. | • Limited dataset diversity reduces effectiveness in adverse conditions (e.g., nighttime, weather variations) [50]. |
| • YOLO-based models demonstrated high-speed real-time detection capabilities (e.g., 71 FPS for AB-DLM) [58]. | • Increased structural complexity (e.g., SE and BiFPN modules) raises computational demands, limiting scalability [58]. |
| • Spatial and channel attention integration in DACNN improved feature discrimination for subtle distractions [59]. | • Short-term performance focus limits long-term adaptability to varied real-world conditions [59]. |
| • SWAM efficiently captures long-distance dependencies and focuses on critical regions using self-attention [60]. | • Trade-off between speed and accuracy due to local receptive field limitations in CNN architectures [60]. |
| • Improved detection of subtle distractions (e.g., phone usage, reaching behind) through multi-attention strategies [59]. | • Reliance on CNN-based backbones constrains generalization across complex real-world scenarios [60]. |
| • Real-time detection capabilities maintained in lightweight models like AB-DLM (15.9 MB) [58]. | • Further training on diverse datasets needed to enhance generalization and robustness [58, 59]. |

*Strengths and Weaknesses associated with Techniques applied to Visual Data Modality*

In summary, for the approaches to distracted driving detection using the visual data modality, the following strengths and weaknesses were identified:

Strengths

1. *High Accuracy and Feature Specificity*: Advanced models such as EfficientDet-D3 and YOLOv8n effectively detect visual distractions by focusing on critical features such as the driver's face, hands, and posture [29, 30, 50].

2. *Transfer Learning Effectiveness:* VGG16 and ResNet architectures, combined with fine-tuning techniques, achieve high accuracy in real-time distraction detection (e.g., 98.5% by VGG16 and ResNet50 combinations) [37, 40].

3. *Real-Time Capabilities:* Lightweight architectures, such as SqueezeNet and MobileNetV3, achieve high frame rates for practical deployment. CNN models integrated with LSTMs also provide temporal flexibility without compromising detection speed [33,46,51].

4. *Effective Spatial-Temporal Integration:* Transformer-based models (e.g., DKT) and hybrid CNN-RNN architectures capture sequential behaviors effectively, improving distraction detection accuracy for dynamic environments [51,52]

5. *Ensemble Model Flexibility:* Ensemble models, such as E2DR and cooperative transfer learning (e.g., combining DenseNet and MobileNet), leverage complementary strengths for improved generalization and robustness [54,55,56].

6. *Attention Mechanism Innovations:* Attention-based models like DACNN and AB-DLM improve focus on relevant features, enabling better generalization across datasets while maintaining computational efficiency [58,59,60].

7. *Improved Generalization Across Datasets:* Domain adaptation techniques like MSDAN and SWAM's attention mechanisms address dataset variability effectively, improving robustness in diverse driving conditions [47,60].

8. *Edge Device Deployment:* Models like YOLOv8n and TensorFlow Lite CNNs are lightweight and compatible with embedded systems, enabling real-world applications in vehicles [34, 50].

Weaknesses

1. *Dataset Limitations:* Models heavily reliant on datasets like State Farm and AUC often lack diversity, limiting their ability to generalize to diverse driving scenarios and complex distractions (e.g., interacting with passengers or eating) [37,54].

2. *Computational Complexity:* Advanced architectures (e.g., AB-DLM, Transformer-based models) require significant computational resources, making deployment on resource-constrained devices challenging [34, 58, 52].

3. *Class Overlap Challenges:* Overlap in visually similar behaviors, such as talking versus eating, can lead to misclassification, particularly in hybrid ensemble and spatial models [51,43].

4. *Incomplete Real-World Testing*: Validation under diverse real-world conditions, such as low-light or adverse weather, is often lacking in studies like CornerNet-Saccade and DACNN [40,60].

5. *Overfitting:* Lightweight and ensemble models are prone to overfitting, particularly when trained on small, imbalanced datasets like State Farm [54,33].

6. *Manual Optimization:* Some methods, such as ADW-KNN, require extensive manual parameter tuning, increasing complexity for practical implementation [28].

7. *Environmental Sensitivity:* Several models struggle in adverse conditions like nighttime driving or poor weather, impacting their robustness [30,50]

8. *Limited Multimodal Integration:* Sole reliance on visual data reduces the models' effectiveness in detecting non-visual distractions (e.g., cognitive or auditory distractions), highlighting the need for multimodal approaches [46,60].

*Sensor (Telemetry) Data*

In the context of driver distraction detection, sensory (telemetry) data provide valuable insights by leveraging real-time information collected from various sensors embedded in or around a vehicle. These sensors can include Inertial Measurement Units (IMUs), GPS, accelerometers, gyroscopes, and other telemetry devices. Unlike purely visual data, sensory data capture *nuanced movements*, *environmental factors*, and *driver interactions* that can be processed to detect patterns indicative of distraction. This section delves into methods that utilize sensory data to enhance the detection of distracted driving behaviors. The approaches covered include machine learning models that analyze sensory inputs and deep learning techniques, which encompass CNNs, temporal and hybrid models, ensemble and multi-task models, and novel architectures. Each subsection will provide an overview of the techniques applied, detailed descriptions of specific models, and an analysis of their strengths and weaknesses.

Figure 26 provides a quick snapshot of the various techniques applied to sensory data modality.

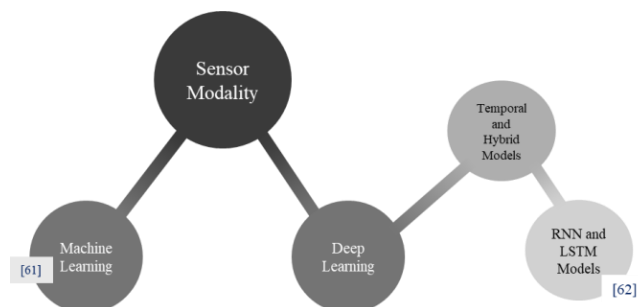

Fig. 26. Summary of Techniques applied to Sensory Data Modality.

*Methods Used*

Machine Learning (ML)

Studies have employed ML techniques by analyzing vehicle and driver metrics. For instance, Richter et al. [62] developed the Phone-Locating-Unit (PLU), a multi-sensor system aimed at detecting and reducing driver distraction caused by mobile phone use. The PLU combines sensory data from *in-cabin cellular activity* and an *IMU*, processed by machine learning algorithms, including neural networks, to accurately differentiate between driver and passenger phone usage. Interestingly, the system operates without needing prior knowledge of the number or locations of smartphones inside the vehicle. Leveraging the IMU's motion tracking, the PLU achieved 98.4% accuracy in challenging scenarios, such as detecting phone use in the driver's right hand or near the gear shift.

However, the study noted *limitations* in *differentiating driver-right-hand* (DRH) phone use from interactions near the gear shift, which remain challenging due to in-cabin signal propagation. The authors suggested future work incorporating a hybrid Received Signal Strength Indicator – Angle of Arrival (RSSI-AoA) algorithm to address these limitations and improve performance without requiring the driver's cooperation. This sensory-driven approach provides a practical solution for real-time driver distraction detection, enhancing safety across various driving contexts.

### Deep Learning (DL)

This section outlines the deep learning techniques employed in processing sensory data for distracted driving detection. It covers approaches including *CNNs, temporal and hybrid models like RNNs* and *LSTMs, ensemble* and *multi-task strategies, attention-based models,* and *novel architectures.* The focus is on highlighting the key frameworks used in the literature and their practical advantages and limitations.

#### *Temporal and Hybrid Models*

##### RNN and LSTM Models

Kouchak and Gaffar [63] proposed a driver distraction detection model using Bidirectional Long Short-Term Memory (Bi-LSTM) and LSTM with an Attention Layer. The system processed simulated driving data, including velocity, steering angle, and lane position, to capture sequential dependencies in driving behavior. The LSTM with attention outperformed the Bi-LSTM, achieving a training Mean Absolute Error (MAE) of 0.85 and testing MAE of 0.96, compared to the Bi-LSTM's MAE of 0.97 (training) and 1.00 (testing). The attention layer effectively improved accuracy by assigning weights to the input sequence, enhancing the model's ability to focus on the most relevant driving data while reducing computational complexity, making it more efficient for real-time distraction detection.

However, the study has several limitations. While the bidirectional LSTM reduced the gap between training and testing errors, its *overall error reduction remained minimal*, necessitating the use of an attention layer to achieve meaningful improvements. The reliance on simulated driving data, rather than real-world conditions, raises concerns about the model's generalizability to actual driving scenarios where environmental factors such as weather, lighting, and road variations come into play. Additionally, the model focuses solely on driving metrics like velocity and lane position, without incorporating multimodal inputs such as physiological or cognitive data, which could provide a more holistic assessment of driver distraction. Although computational efficiency was improved, the scalability of the attention-enhanced LSTM for resource-constrained embedded systems is not fully explored.

#### *Novel Architectures*

Shariff et al. [64] proposed a spiking neural network (SNN) for detecting driver distraction using event-based data from the simulated DMD dataset (v2e). The method effectively handles high temporal resolution data with a *low parameter count*, ensuring computational efficiency and real-time applicability. A *privacy-by-design framework* was integrated to protect sensitive driver information, addressing key privacy concerns. The Spiking-DD networks outperformed existing models in both accuracy and efficiency.

However, the method's reliance on *simulated data* raises concerns about generalizability to real-world scenarios, particularly under varying lighting and environmental conditions. The *real-time capabilities* on hardware accelerators like Intel's Loihi-2 chip remain untested, and further validation on physical hardware is needed to confirm its robustness.

Overall, the study demonstrates the promise of SNNs for distraction detection but highlights the need for extensive testing in real-world conditions and on dedicated hardware platforms. The general overview of the sensing pipeline and spiking-DD neural network architecture are shown in Figures 27 and 28 respectively.

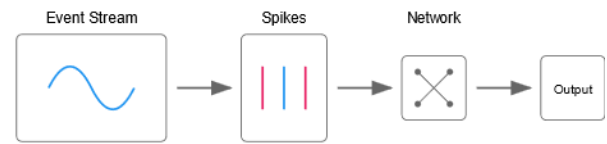

Fig. 27. Overview of sensing pipeline adapted from Shariff et al. [64].

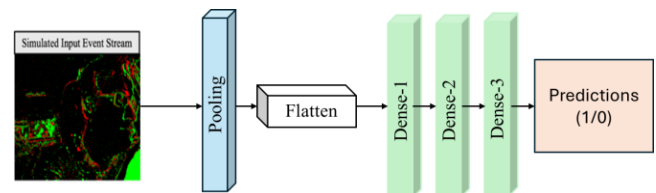

Fig. 28. Proposed spiking-DD neural network. Adapted from Shariff et al. [64].

#### *Strengths and Weaknesses associated with Techniques applied to Sensory Data Modality*

Review of the relevant literature reveals the following strengths and weaknesses related to distracted driving detection using sensory data:

Strengths

1. *High Accuracy with Sensory Fusion:* Multi-sensor systems like PLU achieve *98.4% accuracy* by combining cellular activity and IMU data, even in challenging conditions [62].

2. *Temporal Feature Extraction:* LSTM and Bi-LSTM models capture *sequential driving behaviors*, improving detection of distraction patterns [63].

3. *Attention Mechanism Benefits:* Attention layers enhance LSTM performance by focusing on *relevant input features,* reducing errors and improving efficiency [63].

4. *Computational Efficiency:* Spiking Neural Networks (SNNs) handle *high temporal resolution* data with low parameter counts, suitable for real-time deployment [64].

5. *Privacy-Conscious Design:* Systems like PLU and SNN integrate *privacy-by-design frameworks*, addressing ethical concerns around in-vehicle monitoring [62], [64].

6. *Real-Time Applicability:* Models balance computational efficiency with real-time performance, enabling practical implementation [63], [64].

Weaknesses

1. *Reliance on Simulated Data:* Validation primarily on *simulated datasets* limits generalizability to real-world conditions like varied lighting and weather [62], [63], [64].

2. *Class Ambiguity:* Challenges remain in distinguishing distractions like *driver-right-hand phone use* versus gear interactions due to signal overlap [62].

3. *Lack of Multimodal Integration:* Models rely solely on *telemetry data,* missing physiological or cognitive signals for a holistic assessment [63].

4. *Untested Hardware Deployment:* Real-time performance on dedicated hardware platforms like *Loihi-2* remains unvalidated [64].

5. *Minimal Error Reduction*: Bi-LSTM models show limited overall improvement, despite attention enhancements [63].

6. *Scalability Concerns:* Computational demands of LSTM and SNNs need further evaluation for *embedded systems* [63], [64].

### Combined Modalities

The complex nature of distracted driving behaviors necessitates a more comprehensive approach to its detection and analysis. In recent years, *multimodal analysis* is emerging as a more comprehensive approach to distracted driving detection. This section explores research that employed multimodal sources of data in their approach. In the context of this review, *multimodal* refers strictly to studies that *combined two or more sources of data including visual, sensory, or auditory modalities* for effective driver distraction detection.

Figure 29 provides a quick snapshot of the various techniques applied to sensory data modality.

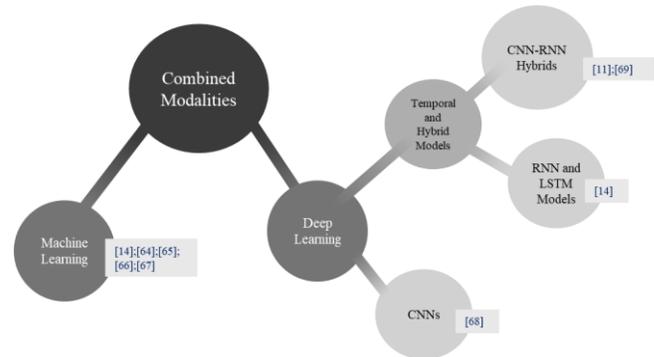

Fig. 29. Techniques Applied to Multimodal Data

*Methods*

Machine Learning (ML)

Several empirical studies have explored the integration of *visual, physiological*, and *sensory data* to enhance multimodal distraction detection using machine learning techniques. Gjoreski et al. [14] demonstrated the effectiveness of combining *video-based features* like facial action units and emotions with *physiological signals* such as heart rate and electrodermal activity, achieving an F1-score of 94% using Extreme Gradient Boosting (XGB). However, limitations such as *label jitter, scenario-specific overfitting*, and *unsatisfactory generalization* reduced the model's robustness. Classical ML methods outperformed deep learning techniques when applied to individual modalities, highlighting the need for further modality optimization.

Misra et al. [65] adopted a similar approach, integrating *eye-tracking data, heart rate variability,* and *vehicle kinematics* using Random Forest and SVM classifiers. The system achieved 90% accuracy in detecting cognitive distractions but suffered from key challenges, including *short data collection durations*, lack of *individual-specific models*, and a *biased participant pool* consisting of young university students. Additionally, the use of a *simple driving simulator* limited ecological validity, underscoring the need for longer, personalized studies and multi-metric evaluations.

Das et al. [66] expanded multimodal integration further by incorporating *thermal signals* alongside visual and physiological data, achieving a 94% F1-score for distraction recognition in a 2-class setting, particularly excelling in detecting *physical* and *frustration distractions*. However, thermal data struggled with *cognitive* and *emotional distractions*, likely due to class imbalance. Visual modalities effectively identified physical distractions but showed *limited scalability* as performance plateaued with increased training samples, while physiological features like the *BVP power spectrum* emerged as critical indicators for distraction detection.

Heenetimulla et al. [67] introduced a real-time FPGA-based system that fused *visual tracking* with *ECG-based physiological monitoring*. The system achieved 93.49% accuracy for mobile phone detection but delivered lower accuracy (78.69%) for driver health monitoring. Key limitations included reliance on *wearable devices* like smartwatches, which may not be universally accessible, and reduced robustness of heart rate variability monitoring compared to mobile usage detection.

Yadawadkar et al. [68] utilized the *SHRP2 Naturalistic Driving Study* dataset, combining *visual data* (e.g., head movements) and *vehicle dynamics* (speed, lane position) to detect distracted and drowsy driving behaviors. While balancing the dataset with SMOTE improved performance, achieving an F1-score of 94%, the study faced challenges with *small training sizes* for LSTM models. Suggested improvements included techniques like *few-shot learning*, *meta-learning*, or *semi-supervised learning* (GANs) to address data limitations. Further recommendations involved testing extended *video sequences (beyond 25 frames)* and integrating *direct features* such as *head pose, gaze,* and *PerClos*, known for their effectiveness in detecting distraction and drowsiness.

Collectively, these studies highlight the strengths of multimodal approaches in improving distraction detection, with *thermal, physiological,* and *visual features* offering complementary benefits. However, challenges such as *class imbalance, generalization issues*, reliance on *wearable devices*, and limited ecological validity emphasize the need for further research to develop scalable, robust, and real-world deployable systems. Figures 30, 31, 32, and 33 represent some of the ML multimodal approaches frameworks.

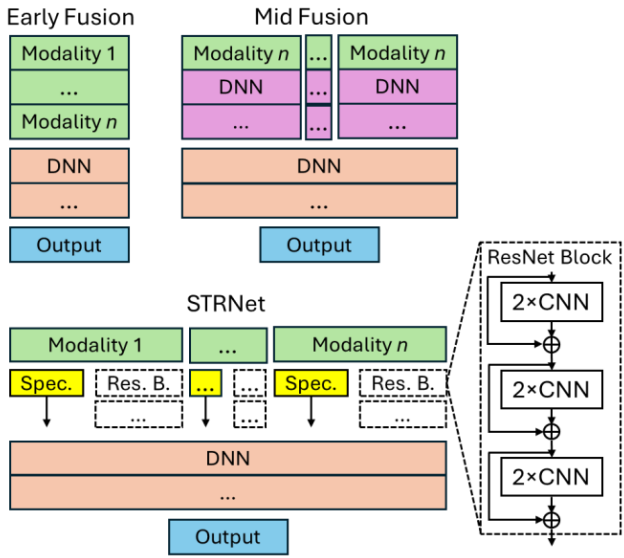

Fig. 30. Fusion approaches illustrated in [14].

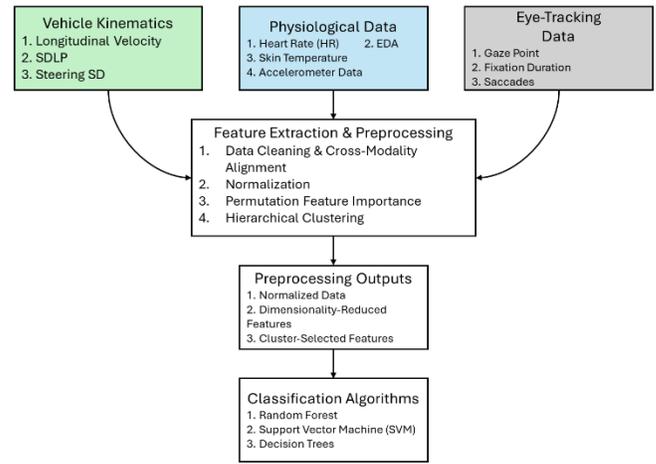

Fig. 31. Multimodal Approach Flowchart adapted from Study [65].

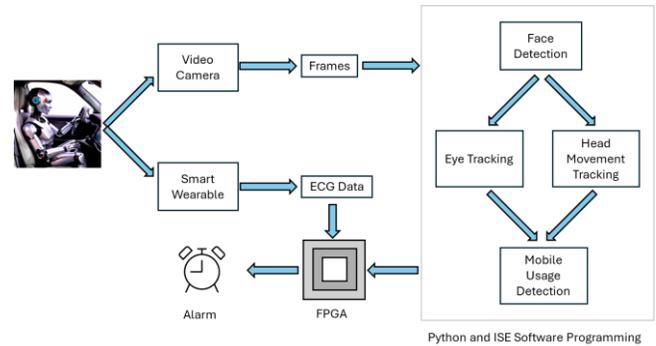

Fig. 32. Block Diagram of Work Formulation adapted from Study [67].

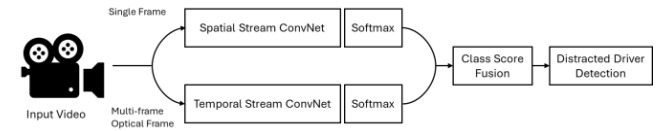

Fig. 33. Model Architecture of CNN video classification adapted from study Yadawadkar et al. [68].

Table X summarizes the strengths and weaknesses identified in the reviewed studies of attention-based models to visual data modalities.

TABLE X
STRENGTHS AND WEAKNESSES OF MACHINE LEARNING
MODELS

| Strengths | Weaknesses |
|---|---|
| • Enhanced Detection: Multimodal integration (visual, physiological, thermal) improves accuracy and complements single-modality limitations [14, 64, 65]. | • Generalization Issues: Simulated and imbalanced datasets limit performance in real-world settings [64, 65, 66]. |
| • Thermal Modality Advantage: Excels in detecting physical and frustration distractions with high F1-scores (e.g., 94%) [65]. | • Dataset Limitations: Short data duration and class imbalances affect cognitive distraction detection [64, 65]. |
| • Real-Time Potential: Lightweight systems (e.g., FPGA-based) show promise for practical deployment [66]. | • Reliance on Wearables: Physiological monitoring systems require devices (e.g., smartwatches), limiting accessibility [66]. |
| • Key Features Identified: Physiological metrics like BVP power spectrum are highly effective for distraction detection [65]. | • Feature Gaps: Missing direct features like head pose, gaze, and PerClos reduce accuracy for subtle distractions [67]. |
| • Improved Data Balance: SMOTE enhances model performance on imbalanced real-world datasets [67]. | • Computational Complexity: Multimodal fusion can increase resource demands, affecting scalability [65, 67]. |

Deep Learning (DL)

*Convolutional Neural Networks (CNNs)*

Martin et al. [69] introduced the *Drive&Act* dataset, a large-scale, multi-modal benchmark designed for driver behavior recognition, featuring 9.6 million frames annotated across 83 behaviors. The dataset combines synchronized RGB, infrared, depth camera data, and 3D skeleton tracking, showcasing the value of integrating visual and sensory inputs for comprehensive behavior analysis. Using CNN architectures such as C3D, P3D ResNet, and I3D, the study demonstrated that I3D outperformed other models, achieving 63.64% accuracy due to its ability to effectively learn spatio-temporal patterns. Accuracy further improved to 69.03% with late fusion techniques, which integrated multiple modalities, highlighting the benefit of combining visual and sensor data to address challenges like poor lighting, partial occlusions, and subtle driver movements.

However, the study noted key limitations. The elaborate nature of annotated actions increased model complexity, while the controlled experimental conditions limited the generalizability of findings to real-world driving scenarios. Additionally, the reliance on hardware-specific sensors, such as Kinect v2, restricts adaptability for broader deployment in commercial vehicle systems. Despite these challenges, the study underscores the potential of combining visual and sensory data in CNN-based frameworks to advance robust driver monitoring systems, ultimately contributing to safer and more responsive vehicular environments.

*Temporal and Hybrid Models*

RNN and LSTM Models

Gjoreski et al. [14] further utilized the Spectro-Temporal ResNet (STRNet) model to process both *visual* and *physiological* signals. This model outperformed the classical ML models for distraction classification, achieving an F1-score of 87% on the test set, demonstrating the effectiveness of end-to-end deep learning for *multimodal signal analysis*. In their study [14], the best DL model was spectro-temporal ResNet.

CNN-RNN Hybrids

Omerustaoglu et al. [11] proposed a multimodal approach that combined visual data (e.g., driver images) with vehicle sensor metrics (e.g., engine RPM, throttle position) to detect driver distractions. Using CNNs (VGG16 and Inception V3) for feature extraction and LSTMs to process fused data, the study explored data-level and prediction-level fusion, achieving a 23% improvement in accuracy, with a maximum accuracy of 85%. However, the dataset used featured images from only a single driver, limiting generalization to diverse demographic groups and driving styles. Expanding the dataset to include varied participants was suggested as a key area for future work.

Anagnostou and Mitianoudis [70] introduced a lightweight ConvGRU model that fused visual data (RGB, infrared) with depth sensor data to monitor driver behaviors. By leveraging depth-wise separable convolutions and GRUs for temporal modeling, the system achieved an F1-score of 0.7205 and AUC of 0.8866 on the DAD dataset, operating at only 24.4% of the computational cost of heavier models. Despite its efficiency, the model's performance lagged behind state-of-the-art architectures like 3D ResNet18 in terms of AUC, and its reliance on controlled datasets limited its generalizability to real-world conditions. Class imbalance handling and the lack of advanced ensemble strategies were identified as areas for improvement, along with addressing reduced accuracy stemming from its lightweight design.

These studies underscore the effectiveness of CNN-RNN hybrids for multimodal distraction detection while highlighting the trade-offs between computational efficiency, accuracy, and generalization capabilities. Figures 34 and 35 are case specific examples of CNN-RNN hybrid models.

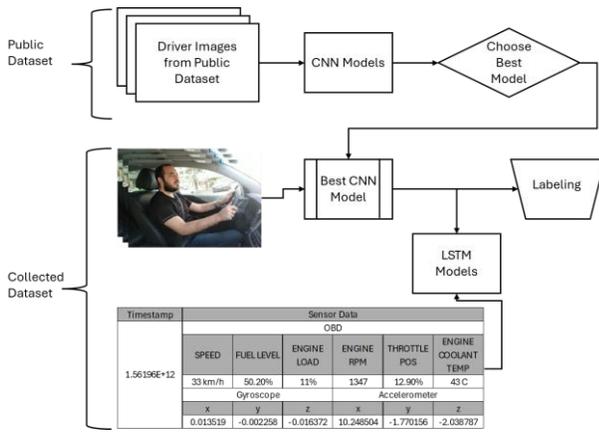

Fig. 34. Overall distracted driver detection system. Adapted from Omerustaoglu et al. [11].

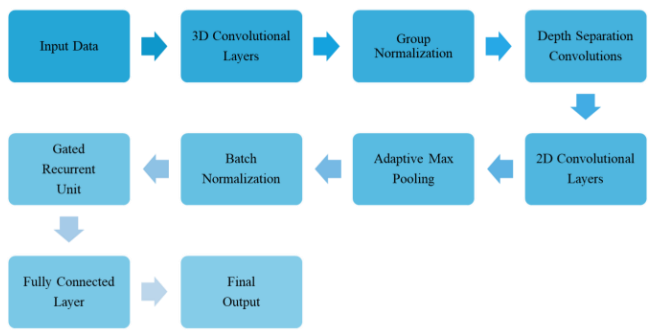

Fig. 35. Lightweight ConvGRU-Based Architecture for Multi-Modal Distracted Driving Detection. Adapted from Anagnostou and Mitianoudis [70].

*Strengths and Weaknesses of Multimodal Approaches to Distracted Driving Detection*

Examination of multimodal approaches to detection of distracted driving revealed the following strengths and weaknesses:

Strengths

1. *High Accuracy Through Multimodal Integration*: Combining visual, thermal, and physiological data improves detection accuracy, achieving F1-scores up to 94% [14, 65].

2. *Temporal Feature Learning*: CNN-LSTM hybrids effectively capture sequential driver behaviors, enhancing performance over standalone models [11, 69].

3. *Computational Efficiency*: Lightweight models like ConvGRU achieve reduced computational costs while maintaining competitive performance [69].

4. *Fusion Flexibility*: Late fusion and data-level fusion techniques improve robustness against challenging conditions like occlusions and lighting variations [11, 68].

5. *Dataset Contributions*: Large-scale benchmarks like Drive&Act provide extensive multi-modal annotations for driver behavior analysis [68].

6. *Real-Time Feasibility*: FPGA-based and lightweight architectures enable real-time implementation for practical applications [66, 69].

Weaknesses

1. *Dataset Limitations*: Controlled conditions and small datasets reduce generalizability to real-world driving scenarios [11, 68, 69].

2. *Class Imbalance*: Imbalanced datasets hinder performance for underrepresented distractions (e.g., cognitive or emotional behaviors) [65, 69].

3. *Hardware Constraints*: Reliance on specific sensors (e.g., Kinect v2, smartwatches) limits scalability in commercial systems [66, 68].

4. *Limited Validation*: Many methods lack comprehensive testing under real-world conditions, such as diverse weather and traffic environments [14, 64, 68].

5. *Overreliance on Single Modalities*: Some studies struggle to effectively integrate all data types, underutilizing key multimodal synergies [65, 67].

6. *Computational Trade-offs*: Lightweight models may lag in accuracy, while advanced systems demand significant computational resources [69, 68].

*Novel Modality*

Raja et al. [71] introduced a Wi-Fi-based distraction detection system leveraging Channel State Information (CSI) to monitor upper body movements, such as head turns and arm gestures, as indicators of distraction. By analyzing phase and subcarrier features extracted from Wi-Fi signals, the system achieved a promising accuracy of 94.5% in classifying distraction behaviors. This radio frequency (RF)-based approach presents a non-intrusive, cost-effective alternative to traditional visual detection systems, making it particularly suitable for real-world vehicle deployment where privacy and scalability are critical.

However, the study highlights key limitations. The system was tested using a single transmitter-receiver pair and lacks validation in scenarios involving multiple passengers, limiting its applicability to real-world conditions. Additionally, the reliance on outdated WLAN cards with modified Linux drivers raises concerns about hardware accessibility and scalability. Future work will require collaboration with device manufacturers to enable seamless access to CSI data, ensuring broader adoption and validation in diverse driving environments. Figure 36 shows the proposed architecture employed in the study.

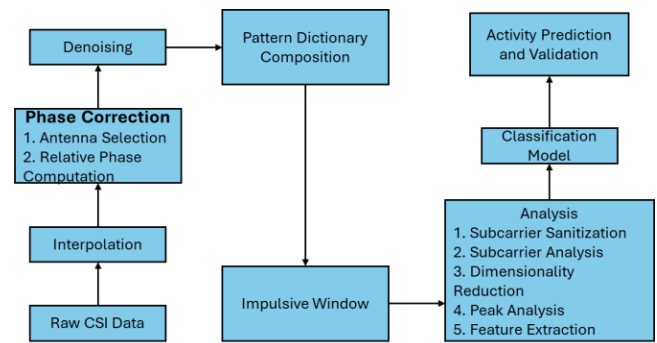

Fig. 36. Proposed System Architecture for Wi-Fi-based distraction. Adapted from Raja et al. [71].

*Private Datasets*

This section provides a comprehensive overview of studies that utilized datasets classified as *private due to their restricted accessibility*. These datasets are integral to understanding advanced applications in distracted driving detection, as they often contain specialized or proprietary data unavailable to the general public. The following subsections mirror the structure used in the Public Datasets section categorizing studies based on the data modality employed. The focus here is exclusively on studies that utilized private datasets, highlighting their unique contribution and limitations to distracted driving detection.

*Visual Data*

As with the public datasets section, this subsection examines methods applied to Visual Data in private datasets, focusing on their unique contributions and limitations.

*Methods Used*

Deep Learning (DL)

As discussed in the Public Datasets section, deep learning techniques are instrumental in analyzing complex data for distracted driving detection. This section focuses on studies utilizing private datasets, highlighting how these methods were adapted to address the specific challenges and opportunities presented by proprietary data.

*Convolutional Neural Networks (CNN-Based Approaches)*

Nur et al. [72] investigated distracted driving behavior classification using a transfer learning approach with a modified VGG16 model. By leveraging visual features including head orientation, hand positioning, and body posture, the researchers achieved an impressive 99.86% accuracy on a private dataset of 10 driving behavior types. While demonstrating significant potential for intelligent transportation systems, the study's limitations include a restricted dataset size and absence of real-world system integration, suggesting critical directions for future research. Figure 37 shows the flowchart of the proposed CNN-based approach.

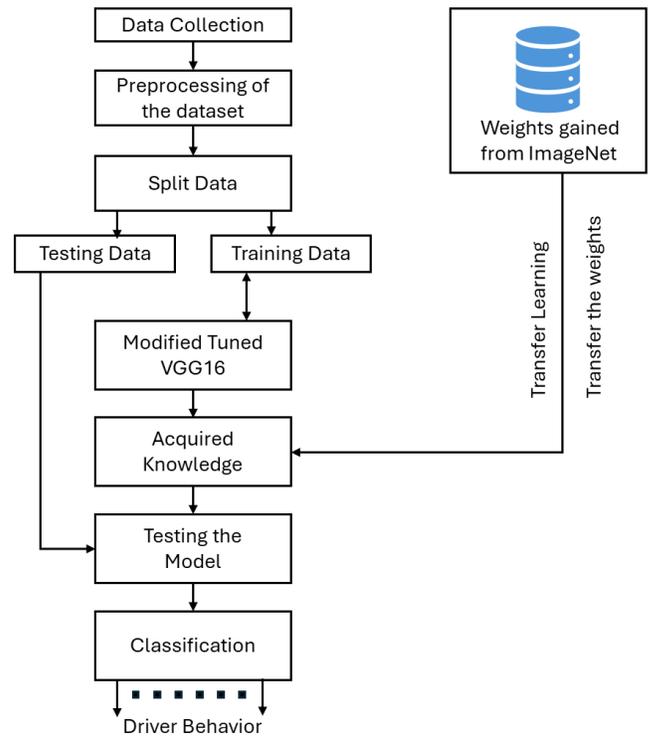

Fig. 37. Flowchart of proposed Transfer learning CNN-based architecture adapted from Nur et al. [72].

*Ensemble and Multi-Task Models*

Draz et al. [73] proposed an ensemble learning approach combining Faster-RCNN for object detection and pose estimation to identify distracted driving behaviors. By utilizing Intersection over Union (IoU) calculations between detected objects and driver pose points, the model achieved 92.2% accuracy on a custom dataset. Despite demonstrating potential for Intelligent Transport Systems, the study revealed limitations including dataset constraints, sensitivity to illumination variations, and significant computational requirements. The approach's narrow focus on hand-held object detection and lack of validation on comprehensive datasets underscore the need for more robust multi-modal distraction detection methodologies. Figure 38 shows the flowchart for the ensemble approach used.

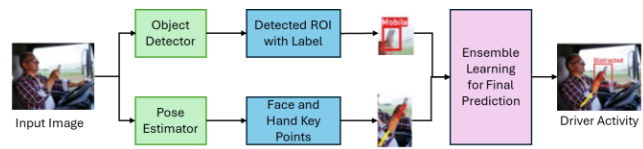

Fig. 38. Flowchart of proposed Transfer learning CNN-based architecture adapted from Draz et al. [73].

*Attention based Models*

Du et al. [74] enhanced YOLOv5 for distracted driving behavior detection by integrating Bidirectional Feature Pyramid Networks (BiFPN) and Convolutional Block Attention Modules (CBAM). Targeting critical visual features like driver actions, hand movements, and body posture, the model achieved 91.6% precision and 89.2% mAP across 14 distraction categories. Despite demonstrating innovative attention-based detection techniques and improved bounding box accuracy through Distance-IoU NMS, the study's limitations include a restricted private dataset and high computational complexity. These constraints underscore the need for more generalized, computationally efficient approaches in real-time driver monitoring systems. The enhanced architecture employed in the study is shown in Figure 39.

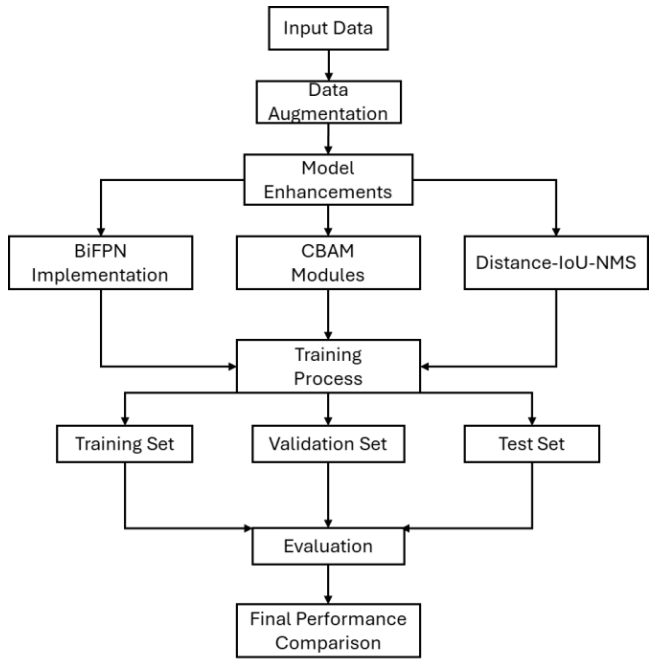

Fig. 39. YOLOv5-DBC: Enhanced Architecture with Path Aggregation and Distance Block Attention adapted from Du et al. [74].

*Strengths and Weaknesses associated with Techniques applied to Visual Data Modality*

Review of literature for approaches using private datasets and the visual modality to identify distracted driving indicate the following strengths and weaknesses:

Strengths
1. *High Accuracy*: Transfer learning with VGG16 [71], Faster-RCNN ensemble approaches, and YOLOv5-based attention mechanisms achieve high accuracy in detecting distracted driving behaviors [73], [74].
2. *Effective Feature Extraction*: Integration of BiFPN and CBAM modules enhances detection of critical visual features such as hand positioning, head orientation, and body posture [72], [73], [74].
3. *Ensemble Flexibility:* Combining object detection with pose estimation (e.g., Faster-RCNN + IoU) improves robustness in identifying driver actions and reducing false positives [73].
4. *Real-Time Potential*: Optimized architectures like YOLOv5 and Faster-RCNN allow for real-time monitoring with high precision and mAP scores [73], [74].

Weaknesses
1. *Dataset Limitations:* Reliance on private or constrained datasets restricts generalizability to diverse real-world scenarios [72], [73], [74].
2. *Computational Complexity*: Advanced modules such as BiFPN and CBAM require significant resources, limiting deployment on low-power devices [73], [74].
3. *Environmental Sensitivity:* Variations in illumination, angles, and backgrounds challenge the models' robustness, particularly for object and pose detection systems [73], [74].
4. *Limited Scope*: Narrow focus on handheld object detection without incorporating multimodal distractions reduces system applicability [73].

*Sensor (Telemetry) Data*

As provided in the public datasets section, sensory (telemetry) data provide valuable insights by leveraging real-time information collected from various sensors embedded in or around a vehicle. This section discusses studies that employed sensory data for distracted driving detection.

Deep Learning (DL)

CNN-RNN Hybrids
Yan et al. [75] developed a novel system for detecting driver distraction using EEG signals during a simulated driving experiment, where distraction was induced using visual and auditory secondary tasks in a 2-back paradigm. The system employed a hybrid model combining EEGNet for feature extraction and LSTM for temporal correlation analysis, achieving a classification accuracy of 71.1% for three-class detection (visual distraction, auditory distraction, and focused state). Also, the model maintained performance with fewer EEG channels (14 instead of 63), improving efficiency without significant accuracy loss.

However, the study faced several limitations. The simplified driving scenario with distractions limited to straightway segments reduces its applicability to real-world, complex environments. Additionally, the overall classification accuracy requires improvement. The study also highlighted the need for a multibranch network to address differences in visual and auditory processing mechanisms, which could further enhance the detection of multimodal distractions.

*Auditory Data*

Auditory data modality is a rarely used modality for driver distraction detection. Zhao et al. [76] introduced a novel driver distraction detection system utilizing wearable acoustic sensors to capture under-skin neck vibrations. By transforming audio signals into Mel-spectrogram features and classifying them using ResNet50V2, the system accurately identified distraction behaviors such as yawning, coughing, sneezing, and talking, achieving an average F1 score of 91.32% on a private dataset containing 8 behaviors. This privacy-friendly approach avoids reliance on visual inputs, providing a non-intrusive alternative for distraction detection.

However, the study identified key limitations. The dataset categories were limited, restricting the system's ability to generalize to more diverse distraction behaviors. Future work is needed to expand the dataset and validate the model's performance across broader and more realistic scenarios. Despite its promise, the system's current scope remains constrained, highlighting the need for further research to explore its applicability in health and physical status monitoring. Zhao et al.'s novel architecture is shown in Figure 40.

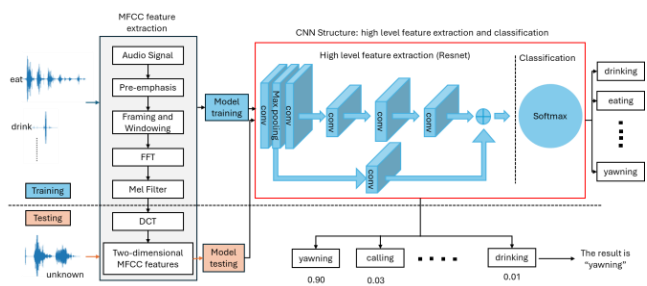

Fig. 40. Overall architecture of the proposed methodology adapted from Zhao et al. [74].

*Comparative Analysis*

The integration of multiple modalities, including visual, sensory, and sometimes auditory data, has emerged as a promising approach for detecting driver distraction. This section compares single-modal and multimodal approaches, highlighting their relative strengths and limitations based on recent studies.

*Cross-modality comparison*

Figure 41 provides a quick snapshot of the observations made from the detailed review conducted in the study.

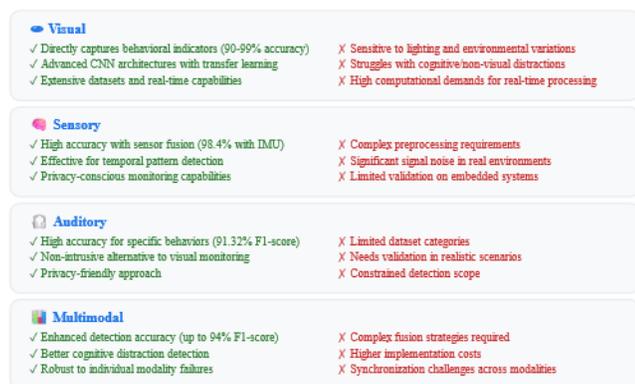

Fig. 41. Cross-modality high level comparison.

## IV. CONCLUSION AND FUTURE WORKS

This comprehensive review has analyzed recent advances in distracted driving detection across visual, sensory, auditory, and multimodal approaches. The analysis reveals several significant developments and critical research gaps that warrant further investigation. In the domain of visual modality, CNN-based approaches have been extensively studied, particularly for detecting manual distractions like phone usage and eating, with transfer learning techniques using established architectures like VGG16 and ResNet demonstrating high accuracy (90-99%) in controlled environments [28, 30, 71]. Significant progress has been made in developing efficient architectures for real-time feature extraction, with lightweight models like MobileNetV3 and YOLOv8n achieving competitive performance while reducing computational demands [34, 50]. The integration of attention modules has been widely explored for improving detection accuracy by focusing on relevant spatial features [58, 59, 60].

However, several critical research gaps persist. Multimodal integration remains underdeveloped, with limited exploration of effective fusion strategies for combining visual, sensory, and auditory data [14, 65] for a comprehensive analysis. Most existing studies rely heavily on controlled datasets with limited environmental variations, highlighting the need for extensive testing under diverse conditions [71, 72, 73]. The temporal aspects of driver behavior, particularly long-term pattern analysis and early distraction detection, remain underexplored [62]. The auditory modality has received significantly less attention compared to visual and sensory approaches, presenting opportunities for investigating speech patterns and ambient sound analysis [75].

Privacy and security considerations in distraction detection systems represent another crucial gap. While some studies have begun addressing privacy-preserving detection methods [61, 63], there remains a need for frameworks that balance monitoring effectiveness with driver privacy. Dataset limitations continue to pose challenges, with a particular scarcity of large-scale, diverse, and publicly available datasets that capture naturalistic driving scenarios [68].

Looking ahead, future research should focus on several key areas. First, robust fusion architectures are needed that can effectively combine multiple modalities while handling real-world challenges like sensor failures and synchronization issues [14, 69]. Second, standardized evaluation frameworks must be created that consider both detection accuracy and practical deployment constraints [46, 47]. Third, lightweight but effective approaches for temporal modeling that can capture long-term behavioral patterns should be investigated [51, 52]. Fourth, privacy-preserving techniques that can protect driver privacy while maintaining detection effectiveness need to be explored [63].

Additionally, there is a pressing need for building comprehensive public datasets that capture diverse driving conditions, demographics, and distraction types [68]. The field would benefit from increased investigation into unsupervised and semi-supervised learning approaches to address the challenge of limited labeled data [67]. Finally, developing adaptive systems that can handle varying environmental conditions and driver behaviors remains crucial [72, 73].

The field of distracted driving detection continues to evolve rapidly, with emerging technologies and methodologies offering new possibilities for enhancing detection accuracy and reliability. By addressing these research gaps and challenges, future work can contribute to developing more robust and practical solutions that effectively enhance road safety. The integration of multiple modalities, coupled with advanced machine learning techniques and privacy-preserving frameworks, holds promise for creating more comprehensive and reliable distraction detection systems [14, 65, 69].

ACKNOWLEDGMENT

The authors would like to thank the Southeast Transportation Workforce (SETWC) for their support.

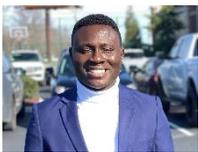
**Anthony Dontoh** is a doctoral candidate in Civil Engineering at the University of Memphis and serves as a graduate research assistant at the Southeast Transportation Workforce and a teaching assistant at the Department. His current research focuses on intelligent transportation systems, road safety, distracted driving detection and multimodal analysis.

He has also contributed high level impact project in advancing gender equality in the transportation sector.

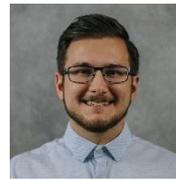
**Logan Sirbaugh**, EI, is a doctoral candidate in Civil Engineering at the University of Memphis and serves as an Instructor in the Herff College of Engineering. He received his B.S. and M.S. degrees in Civil Engineering from the University of Memphis. His research interests include active transportation systems, road ecology, and transportation workforce development, with particular emphasis on bicycle level of traffic stress frameworks and naturalistic driving studies. He is currently a Graduate Research Assistant with the Southeast Transportation Workforce Center, where he has contributed to multiple projects for federal, state, and local transportation agencies.

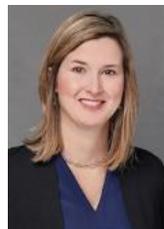
**Dr. Stephanie Ivey** is a Professor of Civil Engineering and Director of the Southeast Transportation Workforce Center in the Herff College of Engineering at the University of Memphis. Her research focuses on transportation planning, operations, and workforce development. She is a member of the Federal Reserve Bank of St. Louis Transportation Industry Council and co-chair of the Institute of Transportation Engineers STEM Committee.

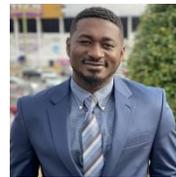
**Dr. Armstrong Aboah** (Member, IEEE) was born in Kumasi, Ghana. He received the B.S. degree in civil engineering from the Kwame Nkrumah University of Science and Technology, in 2017, the M.S. degree in transportation engineering from Tennessee Technological University, in 2019, and the Ph.D. degree in transportation engineering with a focus on computer vision and machine learning from the University of Missouri, Columbia, MO, USA, in 2022. Several of his papers have been published at top venues, including CVPR, NeuIPS, and the Journal of Transportation Engineering. His articles have received more than 243 citations. He has secured competitive research funding from various agencies, including projects on extracting insights from naturalistic driving data and factors influencing transportation network company usage. His research interests include computer vision, transportation sensing, big data analytics, and deep learning, with a focus on intelligent transportation systems. He has authored over 15 peer-reviewed publications in these areas. He is a Leading Researcher of vision-based traffic anomaly detection and has published one of the first papers applying deep learning and decision trees to this problem. His article on real-time multi-class helmet violation detection using few-shot learning techniques has also been impactful. Other novel research contributions include smartphone-based pavement roughness estimation using deep learning, prediction of bus delays across multiple routes, and automated retail checkout using computer vision.